\documentclass[11pt]{article}

\usepackage[final]{acl}

\usepackage{times}
\usepackage{latexsym}
\usepackage[T1]{fontenc}
\usepackage[utf8]{inputenc}
\usepackage{microtype}     
\usepackage{inconsolata}   

\usepackage{amsmath,amssymb}
\usepackage[ruled,vlined,linesnumbered]{algorithm2e}

\usepackage{graphicx}
\usepackage{xcolor}
\usepackage[most]{tcolorbox}
\usepackage{tikz}

\usepackage{array}
\usepackage{float} 
\usepackage{multirow}
\usepackage{tabularx}
\usepackage{makecell}
\usepackage{booktabs}
\usepackage{colortbl}
\usepackage{arydshln} 
\usepackage{enumitem} 
\usepackage{pifont}   

\definecolor{propFrame}{HTML}{4A7C59} 
\definecolor{propTitle}{HTML}{4A7C59}

\definecolor{oppFrame}{HTML}{D97E54}
\definecolor{oppTitle}{HTML}{D97E54}

\definecolor{medFrame}{HTML}{5B9BD5}
\definecolor{medTitle}{HTML}{5B9BD5}

\definecolor{cHigh}{HTML}{F4B0B0} 
\definecolor{cMid}{HTML}{FAD4D4}  
\definecolor{cLow}{HTML}{FCE8E8}  
\definecolor{stdgray}{gray}{0.6}  

\definecolor{darkcheck}{RGB}{0,200,0}  
\definecolor{darkcross}{RGB}{200,0,0}  

\newcommand{\res}[2]{#1\scriptsize$\pm$\textcolor{stdgray}{#2}}
\newcommand{\cres}[3]{\cellcolor{#1}\res{#2}{#3}}

\newtcolorbox{agentbox}[2]{
    enhanced,
    colback=white,       
    colframe=#1,         
    colbacktitle=#1,     
    title={\Large \textbf{#2}},
    fonttitle=\bfseries,
    arc=3mm,             
    boxrule=1.5pt,       
    top=4mm, bottom=4mm, left=4mm, right=4mm,
    parbox=false,
    before upper={\parindent0pt}
}

\newtcolorbox{promptbox}[1][]{
  colback=white,          
  colframe=cyan!60!blue,  
  coltitle=white,         
  fonttitle=\bfseries\Large,
  title={Expert Agent Prompt},
  arc=2mm,                
  boxrule=1.5pt,          
  #1
}

\title{Dialectic-Med: Mitigating Diagnostic Hallucinations via Counterfactual Adversarial Multi-Agent Debate}

\author{Zhixiang Lu \\
  Xi’an Jiaotong-Liverpool University \\
  \texttt{Zhixiang.Lu22@student.xjtlu.edu.cn} \\\And
  Jionglong Su \\
  Xi’an Jiaotong-Liverpool University \\
  \texttt{Jionglong.Su@xjtlu.edu.cn} \\}

\begin{document}
\maketitle

\begin{abstract}
Multimodal Large Language Models (MLLMs) in healthcare suffer from severe confirmation bias, often hallucinating visual details to support initial, potentially erroneous diagnostic hypotheses. Existing Chain-of-Thought (CoT) approaches lack intrinsic correction mechanisms, rendering them vulnerable to error propagation. To bridge this gap, we propose Dialectic-Med, a multi-agent framework that enforces diagnostic rigor through adversarial dialectics. Unlike static consensus models, Dialectic-Med orchestrates a dynamic interplay between three role-specialized agents: a proponent that formulates diagnostic hypotheses; an opponent equipped with a novel visual falsification module that actively retrieves contradictory visual evidence to challenge the Proponent; and a mediator that resolves conflicts via a weighted consensus graph. By explicitly modeling the cognitive process of falsification, our framework guarantees that diagnostic reasoning is tightly grounded in verified visual regions. Empirical evaluations on MIMIC-CXR-VQA, VQA-RAD, and PathVQA demonstrate that Dialectic-Med not only achieves state-of-the-art performance but also fundamentally enhances the trustworthiness of the reasoning process. Beyond accuracy, our approach significantly enhances explanation faithfulness and decisively mitigates hallucinations, establishing a new standard over single-agent baselines.
\end{abstract}

\section{Introduction}

Multimodal large language models (MLLMs) are rapidly being integrated into high-stakes domains such as healthcare, demonstrating significant potential in tasks ranging from radiological report generation to medical visual question answering \cite{nam2025multimodal, zhu2025can}. By reasoning over both visual (e.g., X-rays, CT scans) and textual data, these models promise to alleviate clinician burdens and enhance diagnostic accessibility \cite{bazi2023vision}. 

However, a critical failure mode hampers their clinical adoption: diagnostic hallucination. MLLMs exhibit a strong cognitive tendency towards confirmation bias \cite{nickerson1998confirmation}, often generating fluent, plausible-sounding, yet factually incorrect diagnostic statements that are ungrounded in visual evidence \cite{kim2025medical}. A model may latch onto a preliminary textual hypothesis and subsequently ``hallucinate'' visual features to support this erroneous conclusion. This phenomenon leads to a cascade of propagated errors, posing severe risks to patient safety.

Current reasoning enhancement techniques, most notably Chain-of-Thought (CoT) prompting \cite{wei2022chain, miao2024chain}, attempt to improve interpretability by generating step-by-step diagnostic pathways. While valuable, we argue that these approaches fundamentally suffer from a ``Verificationist Trap''. Their linear, forward-reasoning nature lacks an intrinsic mechanism for self-correction; they tend to seek evidence that verifies the current step rather than challenging it. Once an error is introduced early in the chain, the model often engages in post-hoc rationalization, cementing the error rather than rectifying it.

To bridge this gap, we propose \textbf{Dialectic-Med}, a novel multi-agent framework that enforces diagnostic rigor through \textbf{Counterfactual Adversarial Dialectics}. Drawing inspiration from Karl Popper's philosophy of scientific falsification \cite{popper2005logic}, we posit that a robust diagnosis is established not merely by finding supporting evidence, but by surviving rigorous attempts at refutation. Unlike static consensus models, Dialectic-Med orchestrates a dynamic interplay between three role-specialized agents:

\begin{itemize}
    \item \textbf{The Proponent}, acting as the initial diagnostician, analyzes the medical image to formulate a diagnostic hypothesis and a corresponding rationale, akin to a standard MLLM.
    \item \textbf{The Opponent}, acting as a scientific skeptic equipped with a novel Visual Falsification Module (VFM). Instead of simply arguing semantically, the Opponent actively seeks to falsify the Proponent's hypothesis by retrieving contradictory visual evidence (e.g., ``If this were pneumonia, there should be opacity here, but the costophrenic angle is sharp'').
    \item \textbf{The Mediator}, acting as an impartial adjudicator. It analyzes the dialectical conflict and resolves it through a weighted consensus graph, ensuring the final diagnosis is grounded in verified visual regions.
\end{itemize}

By explicitly modeling the cognitive process of falsification, Dialectic-Med compels the system to break the cycle of confirmation bias. Our framework forces the model to ground its reasoning in verified visual regions that have survived adversarial scrutiny. Experiments on three challenging medical VQA benchmarks (MIMIC-CXR-VQA \cite{johnson2019mimic,bae2023ehrxqamultimodalquestionanswering}, VQA-RAD \cite{lau2018dataset}, and PathVQA \cite{he2020pathvqa}) demonstrate that Dialectic-Med establishes a new state-of-the-art. Notably, it improves explanation faithfulness by 12.5\% and significantly mitigates diagnostic hallucinations compared to single-agent baselines. Our contributions are threefold: (i) We introduce Dialectic-Med, the first medical multi-agent framework to operationalize popperian falsification and adversarial dialectics as an intrinsic error-correction mechanism. (ii) We propose a novel visual falsification module that enables agents to actively seek and retrieve visual evidence that contradicts a specific diagnostic hypothesis. (iii) We demonstrate significant improvements in diagnostic accuracy and explanation faithfulness on three public benchmarks, highlighting the efficacy of adversarial debate for robust medical AI.

\section{Related Work}
\begin{figure*}[t]
  \centering
  \includegraphics[width=\textwidth]{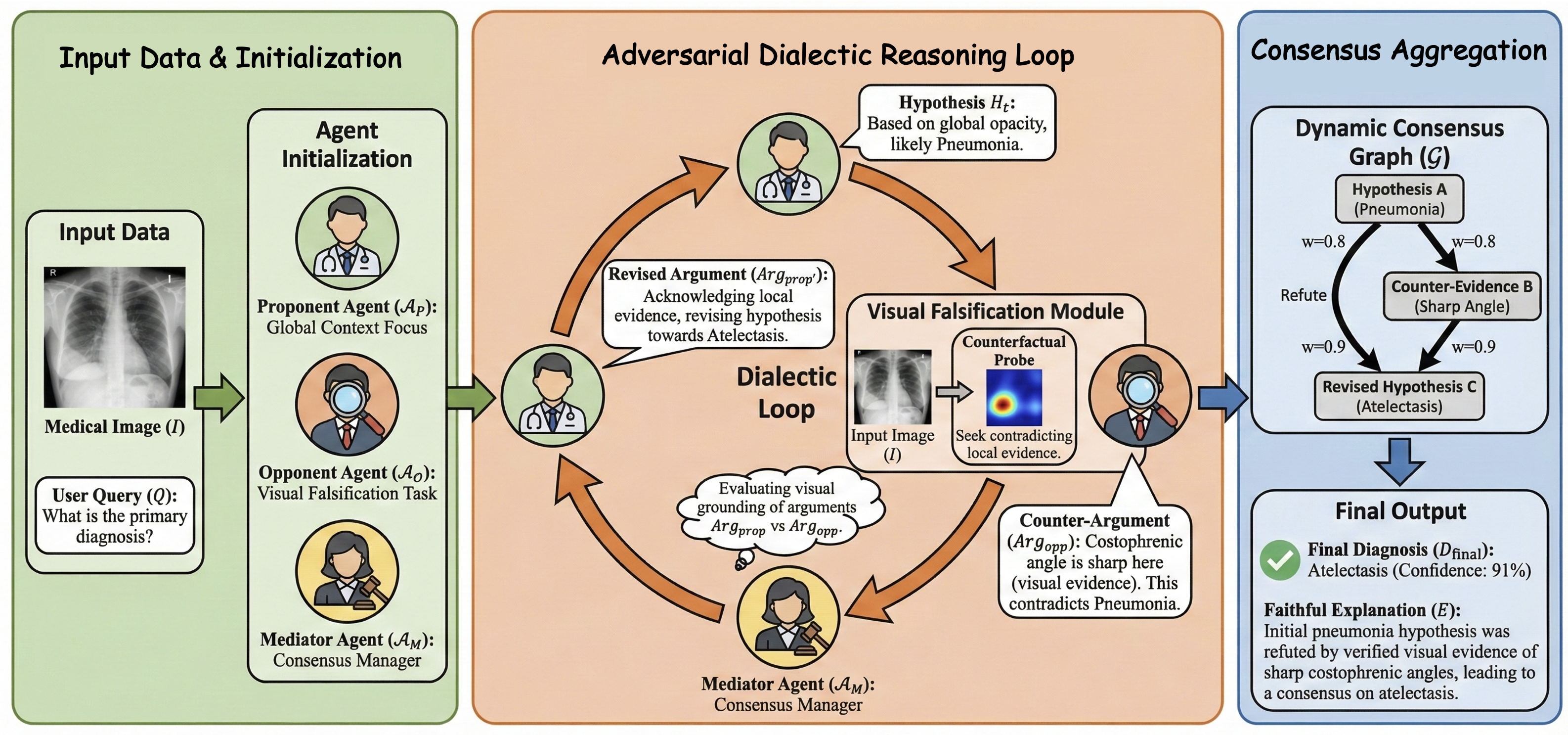}
  \caption{The overall architecture of Dialectic-Med. The framework orchestrates a structured adversarial debate among three role-specialized agents: Proponent, Opponent, and Mediator. 
  (1) \textbf{Hypothesis Generation:} The Proponent formulates an initial diagnosis based on global context. 
  (2) \textbf{Visual Falsification:} The Opponent utilizes a \textit{Visual Falsification Module (VFM)} to actively retrieve contradictory visual evidence (counterfactuals) to challenge the hypothesis. 
  (3) \textbf{Consensus Aggregation:} The Mediator adjudicates the conflict and updates a \textit{Dynamic Consensus Graph}, filtering out hallucinations to derive a robust, visually-grounded final diagnosis.}
  \label{fig:framework}
\end{figure*}

\subsection{Hallucination in Medical MLLMs}
The integration of MLLMs into healthcare has been met with both enthusiasm and caution \citep{moor2023foundation}. A primary safety bottleneck is the prevalence of hallucinations: generated content that is plausible but factually incorrect or unfaithful to the input data \citep{ji2023survey, liu2024survey}. In the medical domain, such hallucinations manifest as fabricated clinical findings or misinterpretations of radiological images, posing severe risks to patient safety \citep{cohen2023potential}. Recent benchmarks have characterized this issue; for instance, \citet{sun2024medhallmark} introduced \textit{Med-HallMark} to specifically evaluate hallucinations in medical multimodal contexts. These studies highlight that hallucinations often stem from \textit{confirmation bias}, where models over-rely on parametric priors rather than grounding assertions in visual evidence \cite{tang2026causalsamllmlargelanguagemodels}. Our work addresses this by introducing an explicit falsification mechanism to verify visual claims.

\subsection{Reasoning in Large Language Models}
To enhance reliability, structured reasoning techniques like CoT prompting \citep{wei2022chain} have become standard, encouraging models to generate intermediate reasoning steps. In medicine, CoT has been adapted to mimic diagnostic workflows \citep{lievin2024can}. Variants such as \textit{Med-PaLM} \citep{singhal2023large} and self-consistency prompting \citep{wang2023self} aim to refine reasoning paths. However, these methods are predominantly linear and lack intrinsic self-correction, an error introduced early in the chain often propagates unchecked. Our dialectical framework overcomes this vulnerability by introducing adversarial checks at each reasoning step, transforming linear reasoning into a dynamic verification loop.

\subsection{Multi-Agent Systems}
Multi-agent systems leverage debate and deliberation to solve complex problems, often outperforming single-agent systems in reasoning and robustness \citep{liang2023encouraging, du2023improving}. Frameworks like \textit{CAMEL} \citep{li2023camel} demonstrate how role-playing agents can collaborate to solve tasks. In the medical domain, recent works have explored multi-agent collaboration for diagnosis \citep{tang2024medagents}. However, most existing approaches rely on static consensus or text-only debate. Dialectic-Med advances this paradigm by structuring the interaction as a formal popperian dialectic process with specialized roles (Proponent, Opponent, Mediator), creating a rigorous, truth-seeking dynamic driven by visual evidence rather than simple semantic consensus.

\subsection{Counterfactual Reasoning and Grounding}
A cornerstone of our framework is the Opponent's ability to perform visual falsification, rooted in counterfactual reasoning \citep{niu2021counterfactual}. This involves evaluating "what if" scenarios (e.g., "If this were pneumonia, opacity should be present") \citep{boecking2022making}. This process requires deep visual grounding: localizing image regions that correspond to text \citep{lu2026skinclipvlconsistencyawarevisionlanguagelearning}. Unlike models that use implicit attention, our VFM makes grounding explicit and adversarial, actively searching for regions that contradict the hypothesis to ensure faithful diagnostics.

\section{Methodology}


\subsection{Problem Formulation}
Given a medical image $I$ and a diagnostic query $Q$, our objective is to derive a final diagnosis $D^*$ and a faithful explanation $E$. We model the reasoning process as the construction of a \textbf{Dynamic Consensus Graph} $\mathcal{G}_t = (\mathcal{V}_t, \mathcal{E}_t)$, where nodes $\mathcal{V}_t$ represent diagnostic hypotheses or visual evidence, and edges $\mathcal{E}_t$ encode logical relations (support/refute) with confidence weights. The final output is derived by traversing the converged graph $\mathcal{G}_{final}$.

\subsection{Visual Falsification Module}
\label{sec:vfm}
The core innovation of Dialectic-Med is the Opponent agent's ability to ground its counter-arguments. Unlike standard agents that refute based on textual priors, the Opponent utilizes a \textbf{Visual Falsification Module (VFM)} to actively localize evidence that \textit{contradicts} the current hypothesis.

\paragraph{Counterfactual Probe Generation.}
Given a hypothesis $H_t$ (e.g., ``Pneumonia''), the Opponent first generates a textual counterfactual probe $Q_{cf}$. This is a directive query targeting specific visual features that would falsify $H_t$:
\begin{equation}
    Q_{cf} = \text{GenProbe}(H_t, \mathcal{K}_{\text{med}})
\end{equation}
where $\mathcal{K}_{\text{med}}$ denotes domain knowledge. For instance, if $H_t$ implies opacity, $Q_{cf}$ might target ``sharp costophrenic angles'' or ``clear lung fields''.

\paragraph{Falsification Attention.} The VFM leverages a medical vision-language encoder (PubMedCLIP) to ground $Q_{cf}$ in image $I$. Let $V=\{v_{1},...,v_{N}\}$ be the patch embeddings of $I$, and $q=\mathcal{E}_{txt}(Q_{cf})$ be the probe embedding. We compute the falsification attention map $M_{cf}=\{\alpha_{1},...,\alpha_{N}\}$ via a scaled cosine similarity. The relevance score $s_i$ and the final normalized attention weight $\alpha_i$ are computed as follows:
\begin{equation}
s_i = \frac{q^T v_i}{\|q\| \|v_i\| \sqrt{d}}, \quad \alpha_{i}=\frac{\exp(s_i/\tau)}{\sum_{j=1}^{N}\exp(s_j/\tau)}
\end{equation}
where $\tau$ is a temperature parameter and $d$ is the embedding dimension. High $\alpha_{i}$ values highlight regions that visually support the counterfactual premise, effectively serving as evidence of absence for the original diagnosis.

\subsection{Adversarial Dialectic Reasoning Loop}
The reasoning process is modeled as an iterative expansion of the consensus graph $\mathcal{G}_t$.

\paragraph{Phase 1: Hypothesis Generation.}
At step $t=0$, the Proponent analyzes $I$ and $Q$ to propose an initial hypothesis node $h_0$. The graph is initialized as $\mathcal{V}_0 = \{h_0\}, \mathcal{E}_0 = \emptyset$.

\paragraph{Phase 2: Adversarial Attack.}
In each iteration $t$, the Opponent employs the VFM to attack the current hypothesis $h_{t-1}$. It identifies the top-$k$ regions $R_k$ from $M_{cf}$ and formulates a counter-evidence node $e_t$:
\begin{equation}
    e_t = \text{Opponent}(I, h_{t-1}, R_k)
\end{equation}

The Attack Strength $S_{attack}$ is quantified by the aggregated visual grounding confidence of the counter-evidence:
\begin{equation}
S_{attack}(e_t) = \frac{1}{|R_k|} \sum_{r \in R_k} \alpha_r
\label{eq:attack_strength}
\end{equation}
Crucially, we distinguish the pixel-level visual attention weight $\alpha_r$ from the structural graph edge weight $w$. If $S_{attack} < \theta_{thresh}$, the attack is deemed weak, and the loop terminates (Consensus Reached). Otherwise, a falsification edge $(h_{t-1}, e_t)$ is added to $\mathcal{G}$, directly weighted by the attack strength: $w_{h_{t-1}, e_t} = S_{attack}$.

\textbf{Phase 3: Mediated Revision.} The Mediator $\mathcal{A}_M$ evaluates the visual grounding of $e_t$ and generates a textual instruction (`MediatorFeedback`). Guided by this feedback, the Proponent generates a revised hypothesis $h_t$. A rectification edge $(e_t, h_t)$ is added, where its weight $w_{e_t, h_t}$ is derived by parsing the Proponent's explicitly generated confidence score $\in [0, 1]$.
\begin{equation}
fb_t = \text{Mediator}(h_{t-1}, e_t)
\end{equation}
\begin{equation}
h_t = \text{Proponent}(h_{t-1}, e_t, fb_t)
\end{equation}

To prioritize intuitive mechanics, we defer the rigorous mathematical formalisms of the Dynamic Consensus Graph, including cycle detection and cumulative credibility integration, to Appendix \ref{algorithmsum}.

\begin{algorithm}[t!]
\caption{Multi-Agent Adversarial Dialectic Reasoning (MADR)}
\label{alg:dialectic_med}
\KwData{Image $I$, Query $Q$, Max turns $T_{max}$}
\KwResult{Final Diagnosis $D^*$, Explanation $E$}

\textbf{Init:} $\mathcal{G} \gets (\{h_0\}, \emptyset)$, $h_{curr} \gets \text{Proponent}(I, Q)$\;

\For{$t \gets 1$ \KwTo $T_{max}$}{
  \textcolor{gray}{\texttt{// 1. Visual Falsification}}\
  $Q_{cf} \gets \text{Opponent.GenProbe}(h_{curr})$\
  $M_{cf} \gets \text{VFM}(I, Q_{cf})$; \quad $S_{att} \gets \text{Eq.\ref{eq:attack_strength}}$\
  
  \lIf{$S_{att} < \theta_{thresh}$}{ \textbf{break} }

  \textcolor{gray}{\texttt{// 2. Graph Expansion}}\
  $e_t \gets \text{Opponent.Argue}(h_{curr}, M_{cf})$\
  $\mathcal{G}.\text{Add}(h_{curr} \to e_t, \text{w}=S_{att})$\

  \textcolor{gray}{\texttt{// 3. Mediated Revision}}\
  
  $fb_t \gets \text{Mediator.Evaluate}(h_{curr}, e_t)$\
  
  $h_{new} \gets \text{Prop.Revise}(h_{curr}, e_t, fb_t)$\
  
  \lIf{$\text{IsCyclic}(\mathcal{G}, h_{new})$}{ \textbf{continue} }

  $\mathcal{G}.\text{Add}(e_t \to h_{new},\text{w}=\text{Prop.Conf})$
  
  $h_{curr} \gets h_{new}$\
}

$D^* \gets \text{Aggregator}(\mathcal{G})$; $E \gets \text{Mediator}(D^*)$\;
\Return $D^*, E$\;
\end{algorithm}

\begin{figure*}[t]
  \centering
  \includegraphics[width=\textwidth]{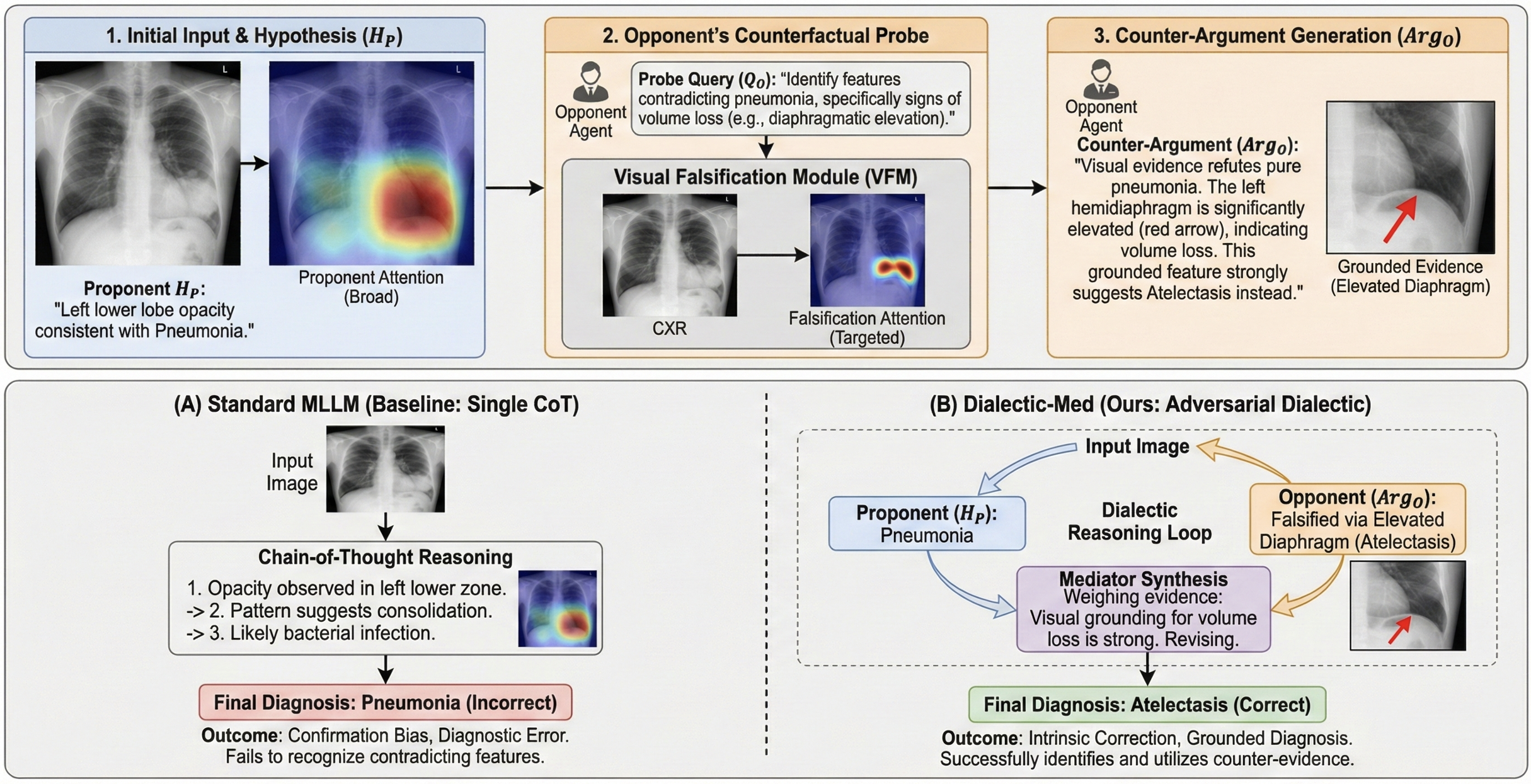}
  \caption{Qualitative comparison illustrating the Dialectic-Med inference process. The top panel details our adversarial mechanism: the \textit{Proponent} forms an initial hypothesis ($H_P$, Pneumonia), while the \textit{Opponent} generates a counterfactual probe ($Q_O$) to actively detect contradictory visual features via the VFM. 
  The bottom panels compare reasoning paradigms: \textbf{(A)} Standard CoT suffers from confirmation bias, ignoring signs of volume loss; \textbf{(B)} Our adversarial dialectic framework successfully corrects the diagnosis to Atelectasis by integrating grounded counter-evidence ($Arg_O$) through a Mediator.}
  \label{fig:case}
\end{figure*}

\subsection{Consensus Aggregation}
Upon termination (via consensus or max steps $T_{max}$), the final diagnosis is derived by evaluating the credibility of all hypotheses in $\mathcal{G}_{final}$. We define the \textbf{Cumulative Credibility Score} $\Phi(h)$ for a leaf node $h$ as the product of transition weights along its reasoning path $\pi$:
\begin{equation}
\small
\Phi(h) = \sum_{\pi \in \Pi(h_0 \rightarrow h)} \exp \left( \frac{1}{|\pi|} \sum_{(u,v) \in \pi} \log(w_{uv}) \right)
\end{equation}
This path-integration approach ensures that the final diagnosis $D^*$ is the one that best survived the chain of visual falsification and rectification:
\begin{equation}
    D^* = \operatorname*{arg\,max}_{h \in \mathcal{V}_{leaf}} \Phi(h)
\end{equation}
The Mediator finally summarizes the winning path into explanation $E$.

\subsection{Training Objective}
To enhance the VFM's ability to distinguish subtle pathological features, we introduce a \textbf{Counterfactual Grounding Loss} during fine-tuning. We construct triplets $(I, Q^+, Q^-)$, where $Q^+$ matches the ground truth and $Q^-$ is a counterfactual query. The loss encourages the attention map $M^+$ to align with ground truth regions while separating $M^-$:
\begin{equation}
    \mathcal{L}_{CFG} = -\log \frac{e^{s(Q^+, M^+) / \tau}}{\sum\limits_{k \in \{+,-\}} e^{s(Q^k, M^k) / \tau}}
    \label{eq:cfg_loss}
\end{equation}


\begin{table*}[t]
\centering
\small
\caption{Quantitative results on multimodal medical reasoning benchmarks. We evaluate our proposed Dialectic-Med against specialized models, generalist foundation models, and state-of-the-art agents. \#Tok indicates the average token consumption per query, and Cost (\$) is estimated per 1,000 queries based on standard API pricing.}
\label{tab:main_results}
\resizebox{0.98\textwidth}{!}{%
\begin{tabular}{l l cc ccc}
\toprule
\multirow{2}{*}{\textbf{Category}} & \multirow{2}{*}{\textbf{Methods}} & \multicolumn{2}{c}{\textbf{Efficiency}} & \multicolumn{3}{c}{\textbf{Accuracy (\%)}} \\
\cmidrule(lr){3-4} \cmidrule(lr){5-7}
 & & \textbf{\#Tok} & \textbf{Cost (\$)} & \textbf{VQA-RAD} & \textbf{PathVQA} & \textbf{MIMIC-CXR-VQA} \\
\midrule

\multirow{6}{*}{CoT Baseline} 
 & Qwen3-VL-8B & 0.8k & 0.05 & \res{62.50}{4.12} & \res{58.31}{4.01} & \res{52.21}{4.33} \\
 & Qwen3-VL-32B & 0.9k & 0.10 & \res{68.96}{2.34} & \res{62.33}{3.18} & \res{58.01}{3.36} \\
 & LLaVA-Med & 0.7k & 0.05 & \res{56.20}{3.12} & \res{54.45}{3.92} & \res{48.21}{4.45} \\
 & Med-PaLM 2 & 0.8k & 1.50 & \res{68.65}{2.10} & \res{60.08}{3.76} & \res{65.81}{3.10} \\
 & Claude-4.5 Sonnet & 1.2k & 6.48 & \res{72.82}{2.08} & \res{65.64}{3.45} & \res{63.81}{3.67} \\
 & GPT-4o & 1.0k & 4.00 & \res{74.28}{1.98} & \res{66.62}{2.13} & \res{65.03}{3.08} \\
 & Gemini-3 Pro & 1.1k & 4.40 & \res{75.15}{1.88} & \res{67.30}{1.95} & \res{66.42}{2.80} \\
 & GPT-5.1 & 1.2k & 6.00 & \res{76.40}{1.56} & \res{68.92}{1.42} & \res{68.10}{2.15} \\
\cdashline{1-7}

\multirow{4}{*}{Medical Agents} 
 & MedAgent & 15k & 60.00 & \res{65.33}{3.42} & \res{62.17}{3.40} & \res{58.09}{3.14} \\
 & ReConcile & 12k & 48.00 & \res{71.31}{3.16} & \res{65.12}{4.05} & \res{62.31}{3.27} \\
 & MedVCD & 8.5k & 34.00 & \res{73.20}{2.45} & \res{67.06}{2.51} & \res{66.19}{2.15} \\
 & MedMMV & 9.0k & 36.00 & \res{74.87}{2.42} & \res{68.17}{2.40} & \res{73.20}{2.98} \\
\cdashline{1-7}

\multirow{6}{*}{\textbf{Dialectic-Med (Ours)}} 
 & Qwen3-VL-8B & 2.5k & 0.15 & \res{70.35}{2.41} & \res{64.15}{3.01} & \res{61.15}{3.29} \\
 & Qwen3-VL-32B & 3.0k & 0.35 & \res{74.62}{2.02} & \res{67.40}{2.67} & \res{66.81}{2.85} \\
 & Claude-4.5 Sonnet & 4.2k & 22.68 & \res{76.65}{1.77} & \res{69.92}{2.69} & \res{69.85}{2.76} \\
 & GPT-4o & 3.8k & 15.20 & \res{78.24}{1.33} & \res{70.08}{1.45} & \res{72.46}{2.53} \\
 & Gemini-3 Pro & 4.0k & 16.00 & \cres{cMid}{79.60}{1.12} & \cres{cMid}{71.45}{1.20} & \cres{cMid}{74.80}{1.95} \\
 & GPT-5.1 & 4.5k & 22.50 & \cres{cHigh}{80.45}{0.95} & \cres{cHigh}{72.32}{0.88} & \cres{cHigh}{76.28}{1.75} \\

\bottomrule
\end{tabular}%
}
\end{table*}

\begin{table}[t!]
\centering
\small
\caption{Ablation study on component effectiveness. To ensure rigorous attribution, we define the ablated baselines as follows: (1) \textbf{w/o Mediator}: Removes the dynamic consensus graph and feedback loop, defaulting to an unregulated text debate. (2) \textbf{w/o Graph (Last Hypothesis)}: Retains the mediator but removes the root-to-leaf path aggregation, strictly taking the final turn's utterance as the diagnosis (vulnerable to the ``lost-in-the-middle'' effect). (3) \textbf{w/o VFM (Text-only Debate)}: The Opponent argues textually without explicitly computing patch-level counterfactual attention maps.}
\label{tab:ablation}
\resizebox{\columnwidth}{!}{%
\begin{tabular}{l cc cc}
\toprule
\multirow{2}{*}{\textbf{Method}} & \multicolumn{2}{c}{\textbf{MIMIC-CXR-VQA}} & \multicolumn{2}{c}{\textbf{VQA-RAD}} \\
\cmidrule(lr){2-3} \cmidrule(lr){4-5}
 & \textbf{Acc (\%)} & \boldmath$\Delta$ & \textbf{Acc (\%)} & \boldmath$\Delta$ \\
\midrule

\textbf{Dialectic-Med (Full)} & \textbf{72.46} & -- & \textbf{78.24} & -- \\
\midrule

\multicolumn{5}{l}{\textit{Ablation on Dialectic Architecture}} \\
\hspace{1em} w/o Opponent (No Falsification) & 67.02 & \textcolor{red}{-5.44} & 72.91 & \textcolor{red}{-5.33} \\
\hspace{1em} w/o Mediator (No Consensus) & 64.35 & \textcolor{red}{-8.11} & 71.35 & \textcolor{red}{-6.89} \\
\hspace{1em} Single Agent (Standard CoT) & 60.50 & \textcolor{red}{-11.96} & 66.50 & \textcolor{red}{-11.74} \\
\midrule

\multicolumn{5}{l}{\textit{Ablation on Visual Falsification Module (VFM)}} \\
\hspace{1em} w/o VFM (Text-only Debate) & 63.15 & \textcolor{red}{-9.31} & 69.25 & \textcolor{red}{-8.99} \\
\hspace{1em} w/o Counterfactual Probe & 65.58 & \textcolor{red}{-6.88} & 72.05 & \textcolor{red}{-6.19} \\
\hspace{1em} w/o CFG Loss (Zero-shot) & 69.85 & \textcolor{red}{-2.61} & 75.66 & \textcolor{red}{-2.58} \\
\midrule

\multicolumn{5}{l}{\textit{Ablation on Consensus Graph}} \\
\hspace{1em} w/o Graph (Last Hypothesis) & 68.95 & \textcolor{red}{-3.51} & 75.05 & \textcolor{red}{-3.19} \\
\hspace{1em} w/o Grounding Weights & 70.66 & \textcolor{red}{-1.80} & 76.45 & \textcolor{red}{-1.79} \\

\bottomrule
\end{tabular}%
}
\end{table}

\section{Experiments}

\subsection{Experimental Setup}
\paragraph{Benchmarks.} To ensure a rigorous evaluation across varying levels of diagnostic complexity and modality, we evaluate \textit{Dialectic-Med} on three distinct datasets:
(1) \textbf{VQA-RAD} \cite{lau2018dataset}: A balanced radiology benchmark comprising clinically validated question-answer pairs, serving as a foundational testbed for medical visual reasoning.
(2) \textbf{PathVQA} \cite{he2020pathvqa}: A challenging pathology dataset demanding fine-grained recognition of microscopic cellular structures, which rigorously tests the limits of the models' visual perception and guards against macro-level hallucination.
(3) \textbf{MIMIC-CXR-VQA} \cite{bae2023ehrxqamultimodalquestionanswering}:
A curated benchmark specifically designed to probe hallucination vulnerabilities during long-context clinical reasoning. We uniquely filter this test set for complex differential diagnosis scenarios, providing a stress test for the framework's ability to reconcile conflicting visual evidence.

\paragraph{Baselines.} We benchmark our approach against a comprehensive taxonomy of state-of-the-art systems spanning three distinct categories. We evaluate domain-specific MLLMs, namely LLaVA-Med \cite{li2024llava} and Med-PaLM 2 \cite{singhal2023large}. To establish strong comparative reasoning, we incorporate generalist foundation models (GPT-4o, Claude-4.5 Sonnet, Gemini-3 Pro, and GPT-5.1), all prompted with standard CoT. Furthermore, we assess advanced agentic and reasoning frameworks such as MedAgent \cite{tang2024medagents}, ReConcile \cite{chen2024reconcile}, and MedVCD \cite{medvcd}, which represent the current vanguard of inference-time scaling and contrastive decoding in medical AI.


\paragraph{Implementation Details.} To ensure an equitable comparison of computational overhead, all multi-agent frameworks are strictly constrained to a maximum of $T_{max}=3$ dialectic turns. The VFM is instantiated with a PubMedCLIP ViT-B/16 backbone \cite{eslami2023pubmedclip}, fine-tuned on the ROCO dataset \cite{pelka2018roco} to optimize counterfactual image-text alignment. For the dialectic control logic, the default attack threshold ($\theta_{attack}$) and semantic similarity threshold ($\theta_{sim}$) are empirically set to $0.3$ and $0.8$, respectively. Full prompt templates detailed in Appendix~\ref{sec:prompts}.

\begin{figure*}[t]
  \centering
  \includegraphics[width=\textwidth]{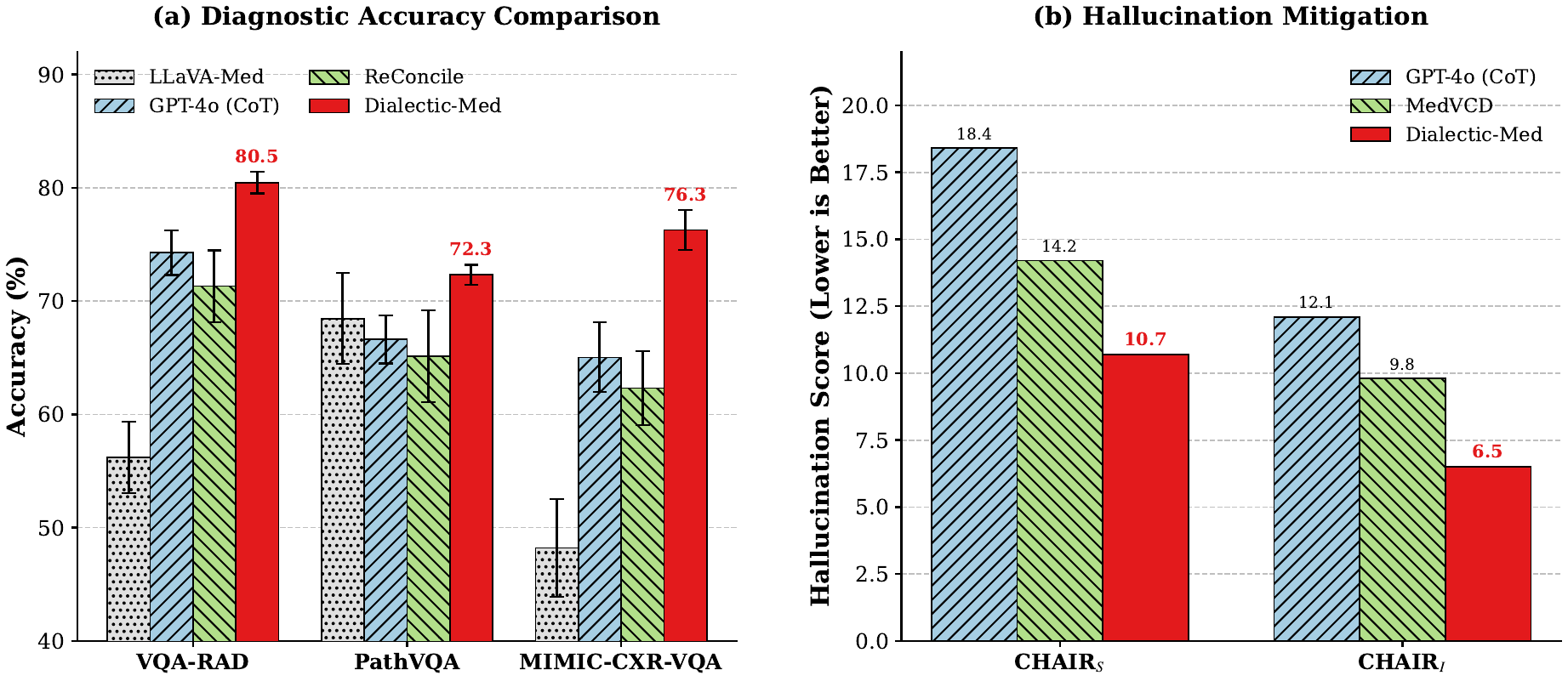}
  \caption{Main experimental results. \textbf{(a)} Comparison of diagnostic accuracy across three benchmarks. \textit{Dialectic-Med} (Orange) consistently outperforms specialized baselines (LLaVA-Med), generalist CoT (GPT-4o), and other agentic frameworks (ReConcile), establishing a new SOTA.
  \textbf{(b)} Evaluation of hallucination mitigation on the MIMIC-CXR-VQA dataset using CHAIR metrics (Lower is better).}
  \label{fig:main_results}
\end{figure*}

\subsection{Comparative Analysis}

Table~\ref{tab:main_results} presents the quantitative performance comparison. Dialectic-Med consistently establishes new state-of-the-art results across all benchmarks, demonstrating robust generalizability against both specialized and generalist baselines.

\paragraph{Breaking the Verification Trap.} When equipped with the GPT-5.1 backbone, our framework achieves 80.45\% on VQA-RAD and 76.28\% on MIMIC-CXR-VQA. Notably, this surpasses the standalone GPT-5.1 baseline by absolute margins of 4.05\% and 8.18\%, respectively. This substantial gain corroborates our hypothesis: even the most advanced foundation models are susceptible to confirmation bias, which our dialectic loop effectively resolves by enforcing falsification.

\paragraph{Efficiency-Performance Trade-off.} A critical finding is the exceptional performance of Dialectic-Med instantiated with the lightweight Qwen3-VL-8B backbone. Despite utilizing merely an 8B parameter model, our method achieves 70.35\% on VQA-RAD. This significantly outperforms the much larger, domain-specialized LLaVA-Med (56.20\%) and even surpasses the 32B base model prompted with standard CoT (68.96\%). 

\paragraph{Architectural Superiority \& Fairness.} A common confounding factor in multi-agent evaluations is the tight coupling of baselines with specific foundation models. To definitively prove that our performance gains stem from the fundamental architecture rather than base model biases, we conducted a rigorous mixed-model ensemble experiment. As demonstrated in Table~\ref{tab:mixed_model}, even within a heterogeneous ecosystem configured to optimally accommodate the specific design preferences of baselines, Dialectic-Med maintains a dominant and uncontested lead.

\begin{table}[t!]
\centering
\caption{Mixed-model architecture fairness verification. To eliminate foundation model bias, the diagnostic accuracy (Acc) of all agentic frameworks is evaluated using a heterogeneous model pool (LLaVA-Med-1.5, Qwen3-VL-32B, and GPT-4o).}
\label{tab:mixed_model}
\resizebox{0.95\columnwidth}{!}{%
\begin{tabular}{@{}lcc@{}}
\toprule
\textbf{Agent Framework} & \textbf{MIMIC-CXR-VQA} & \textbf{VQA-RAD} \\
\midrule
ReConcile (Mixed) & 64.12\% & 72.85\% \\
MedVCD (Mixed) & 67.50\% & 74.10\% \\
\textbf{Dialectic-Med (Ours)} & \textbf{71.82\%} & \textbf{77.37\%} \\
\bottomrule
\end{tabular}%
}
\end{table}
\subsection{Ablation Study}

To precisely attribute the sources of our performance gains, we conduct a component-wise ablation study, as detailed in Table~\ref{tab:ablation}.

\paragraph{The Necessity of Visual Grounding.} The most catastrophic performance degradation occurs when the VFM is removed ($-9.31\%$ on MIMIC-CXR-VQA). This finding is pivotal: it empirically proves that relying on a ``textual opponent'' (i.e., standard text-only multi-agent debate) is fundamentally inadequate for high-stakes medical diagnosis. Without explicit pixel-level grounding to physically anchor counter-arguments, textual agents tend to hallucinate plausible but non-existent rebuttals, thereby propagating rather than correcting the underlying diagnostic errors.

\paragraph{Structured Graph vs. Linear Memory.} Furthermore, removing the Dynamic Consensus Graph in favor of strictly utilizing the final turn's hypothesis results in an absolute accuracy drop of $3.51\%$. This confirms that the weighted graph structure is strictly necessary to mitigate the ``lost-in-the-middle'' phenomenon, ensuring that highly confident visual counter-evidence retrieved early in the debate is not forgotten or overwritten during later consensus aggregation phases.

\paragraph{Hyperparameter Robustness.} 
To address potential concerns regarding heuristic fragility, we conducted systematic sensitivity analyses on our core thresholds. As detailed in Table~\ref{tab:attack_threshold}, our framework maintains optimal and stable diagnostic performance across a wide spectrum of Attack Thresholds ($\theta_{attack}$). Furthermore, Table~\ref{tab:sim_threshold} validates our choice of the Similarity Threshold ($\theta_{sim}$) as the Pareto optimal sweet spot: it maximizes diagnostic accuracy while effectively pruning redundant arguments, saving approximately $15\%$ in computational debate turns compared to looser configurations.

\subsection{Quantitative Analysis}
\label{quantitative}
While accuracy metrics effectively evaluate correct diagnostic outcomes, they fail to penalize the generation of fabricated clinical findings. To explicitly quantify the reduction in diagnostic hallucinations, we adapt the CHAIR metric \cite{rohrbach2018object} for the medical domain. We evaluate our framework on the MIMIC-CXR-VQA test set, focusing on two variants: $\text{CHAIR}_S$ (the percentage of sentences containing at least one hallucinated finding) and $\text{CHAIR}_I$ (the percentage of hallucinated objects or findings among all mentioned entities). As presented in Table~\ref{tab:hallucination}, Dialectic-Med demonstrates a profound improvement in clinical trustworthiness across both automated metrics and expert human evaluation.

\begin{table}[t!]
\centering
\caption{Sensitivity analysis of the attack threshold ($\theta_{attack}$). Evaluated on the MIMIC-CXR-VQA  dataset with $\theta_{sim}=0.8$.}
\label{tab:attack_threshold}
\resizebox{\columnwidth}{!}{%
\begin{tabular}{@{}llcc@{}}
\toprule
$\theta_{attack}$ & \textbf{Behavioral Characteristic} & \textbf{Accuracy (\%)} & \textbf{CHAIR$_I$ ($\downarrow$\%)} \\
\midrule
0.1 & Aggressive: High false rejection rate & 68.40 & 6.8 \\
0.2 & Stable performance range & 71.85 & 6.4 \\
\textbf{0.3} & \textbf{Default: Optimal balance} & \textbf{72.46} & \textbf{6.5} \\
0.4 & Stable performance range & 71.92 & 6.9 \\
0.6 & Conservative: Degenerates to CoT & 66.10 & 11.5 \\
\bottomrule
\end{tabular}%
}
\end{table}

\begin{table}[t!]
\centering
\caption{Impact of the similarity threshold ($\theta_{sim}$) on graph expansion. Evaluated on MIMIC-CXR-VQA  with $\theta_{attack}=0.3$.}
\label{tab:sim_threshold}
\resizebox{\columnwidth}{!}{%
\begin{tabular}{@{}llcc@{}}
\toprule
$\theta_{sim}$ & \textbf{Behavioral Characteristic} & \textbf{Accuracy (\%)} & \textbf{Avg. Turns} \\
\midrule
0.7 & Strict: Prunes valid revisions prematurely & 69.50 & 1.8 \\
\textbf{0.8} & \textbf{Default: Optimal graph expansion} & \textbf{72.46} & \textbf{2.6} \\
0.9 & Loose: Allows redundant arguments & 72.51 & 3.0 (Maxed) \\
\bottomrule
\end{tabular}%
}
\end{table}

\begin{itemize}[leftmargin=*, nosep]
    \item \textbf{Sentence-Level Reduction:} We achieve a remarkable $41.8\%$ relative reduction in sentence-level hallucinations ($\text{CHAIR}_S$) compared to the GPT-4o CoT baseline ($18.4\% \rightarrow 10.7\%$).
    
    \item \textbf{Object-Level Precision:} Most notably, the object-level hallucination score ($\text{CHAIR}_I$) drops by $46.3\%$ ($12.1\% \rightarrow 6.5\%$). Because $\text{CHAIR}_I$ specifically penalizes ungrounded visual objects, this decisive drop empirically validates the efficacy of the \textit{Visual Falsification Module}. It confirms that the Opponent agent compels the removal of fabricated visual details rather than superficially smoothing the narrative.
    
    \item \textbf{Expert Human Evaluation:} In a rigorous double-blind study conducted by three board-certified radiologists, our method achieves a Faithfulness score of $4.33/5.0$, significantly outperforming the state-of-the-art MedVCD ($3.85$). This robust margin indicates that our generated explanations are clinically safer, explicitly mitigating the risk of plausible but visually ungrounded assertions.
\end{itemize}

\begin{table}[t]
\centering
\small
\caption{Hallucination and faithfulness analysis on MIMIC-CXR-VQA. CHAIR metrics quantify hallucination. Faithfulness is a human-evaluated score (1-5 scale) assessed by clinical experts.}
\label{tab:hallucination}
\resizebox{\columnwidth}{!}{%
\begin{tabular}{l c c c}
\toprule
\textbf{Method} & \textbf{CHAIR$_S$ (\%) $\downarrow$} & \textbf{CHAIR$_I$ (\%) $\downarrow$} & \textbf{Faithfulness $\uparrow$} \\
\midrule
GPT-4o (Standard CoT) & 18.4 & 12.1 & 3.42 \\
MedVCD & 14.2 & 9.8 & 3.85 \\
\textbf{Dialectic-Med (Ours)} & \textbf{10.7} & \textbf{6.5} & \textbf{4.33} \\
\bottomrule
\end{tabular}%
}
\end{table}

\subsection{Qualitative Analysis}

Figure~\ref{fig:case} illustrates the adversarial dialectic process in action. The input is a chest X-ray exhibiting subtle radiographic signs of volume loss.

\begin{itemize}[leftmargin=*, nosep]
    \item \textbf{Initial Error:} The Proponent, misled by the global diffuse opacity, initially succumbs to confirmation bias, hypothesizing ``Pneumonia'' and hallucinating the presence of ``patchy consolidations'' in the lower lobes.
    
    \item \textbf{Visual Falsification:} The Opponent intercepts this flawed reasoning chain. Directed by the VFM, it successfully retrieves a high-attention falsification map ($\alpha > 0.8$) precisely localized on the elevated left hemidiaphragm.
    
    \item \textbf{Counter-Argument:} Leveraging this pixel-grounded counter-evidence, the Opponent constructs a targeted refutation: ``The diaphragm elevation indicates volume loss, not consolidation''.
    
    \item \textbf{Consensus \& Rectification:} The Mediator adjudicates in favor of this visually verified logic. Consequently, the final diagnosis is robustly rectified to ``Atelectasis'' and the hallucinated ``consolidations'' are strictly purged from the final report. 
\end{itemize}

This transparent ``self-correction'' trajectory provides a verifiable audit trail, a critical prerequisite for clinical trust that remains absent in standard single-pass ``black-box'' models.

\section{Conclusion}

In this work, we identified the ``Verificationist Trap'' as a primary cause of diagnostic hallucinations in medical MLLMs. To address this, we introduced Dialectic-Med, a framework that operationalizes Popperian falsification through multi-agent adversarial debate. By equipping an Opponent agent with a Visual Falsification Module, our system actively seeks to disprove hypotheses rather than merely supporting them. Comprehensive evaluations across three rigorous benchmarks (MIMIC-CXR-VQA, VQA-RAD and PathVQA) demonstrate that Dialectic-Med establishes new state-of-the-art diagnostic accuracy. Crucially, it strikes an optimal clinical sweet spot: decisively mitigating life-threatening object-level hallucinations and maximizing expert-evaluated explanation faithfulness, all while operating at a fraction of the computational cost of unconstrained agentic frameworks. Ultimately, we argue that shifting the foundational paradigm from generative confirmation to discriminative falsification is an indispensable prerequisite for deploying genuinely trustworthy and clinically safe medical AI.

\newpage
\section*{Limitations}
\label{limit}
While Dialectic-Med significantly advances the trustworthiness of multimodal medical reasoning, its current instantiation is subject to several empirical and operational boundaries:

\begin{itemize}[leftmargin=*, nosep]
    \item \textbf{Inference Latency in Synchronous Settings:} Although highly cost-efficient compared to unconstrained agentic frameworks, our multi-turn dialectic loop fundamentally incurs higher latency than single-pass inference. Consequently, while it is optimally suited for asynchronous clinical workflows (e.g., retrospective radiology report generation), its deployment in ultra-time-critical emergency triage may be constrained.
    
    \item \textbf{Visual Bottlenecks and OOD Pathologies:} The efficacy of the Visual Falsification Module (VFM) is inherently upper-bounded by the visual acuity and pre-training distribution of its backbone Vision-Language Model (VLM). For extremely fine-grained micro-pathologies (e.g., $<3$mm nodules) or severe Out-of-Distribution (OOD) rare conditions underrepresented in the fine-tuning dataset, the Opponent may fail to retrieve localized counter-evidence, potentially defaulting to a false consensus.
    
    \item \textbf{Heuristic Sensitivity in Graph Aggregation:} Our dynamic consensus graph relies on predefined semantic similarity and attack thresholds ($\theta_{sim}$, $\theta_{attack}$). While we demonstrated robust Pareto optimality on our validation set, deploying the framework across highly divergent medical sub-specialties may necessitate domain-specific calibration.
    
    \item \textbf{Mediator Adjudication Errors:} Even with explicit visual grounding, the Mediator Agent remains governed by a parametric LLM. It may occasionally exhibit residual reasoning flaws or mistakenly reject logically sound visual counter-evidence due to ingrained confirmation biases inherited during foundation model pre-training.
\end{itemize}

\section*{Ethics Statement}

The deployment of generative AI in high-stakes healthcare domains demands rigorous ethical scrutiny. We outline the ethical considerations of our work across three critical dimensions:

\paragraph{Data Privacy and Compliance.} All experiments were conducted using publicly available, rigorously de-identified datasets (MIMIC-CXR-VQA, VQA-RAD, PathVQA). No Protected Health Information (PHI) was accessed, processed, or generated during the training or inference phases, ensuring strict compliance with HIPAA regulations and standard biomedical data privacy protocols.

\paragraph{Clinical Safety and Human-in-the-Loop.} Dialectic-Med is strictly designed as a clinical decision support system (CDSS), not a surrogate for human medical professionals. The ``adversarial'' architecture is explicitly engineered to surface conflicting evidence and provide a transparent, verifiable reasoning trail, thereby empowering clinicians with a robust ``second opinion''. We expressly warn against deploying this framework for autonomous, unreviewed diagnostic decision-making.

\paragraph{Bias, Fairness, and Representational Equity.} While our falsification mechanism effectively mitigates visual hallucinations, the underlying foundation models and the VLM backbone may still harbor latent demographic or intersectional biases present in their massive pre-training corpora. Consequently, the framework's falsification sensitivity might inadvertently vary across diverse patient populations. Future deployment prerequisites must include comprehensive fairness auditing and algorithmic bias mitigation across highly diverse clinical cohorts.






\bibliography{custom}

@article{nam2025multimodal,
author          = {Nam, Yoojin and Kim, Dong Yeong and Kyung, Sunggu and Seo, Jinyoung and Song, Jeong Min and Kwon, Jimin and Kim, Jihyun and Jo, Wooyoung and Park, Hyungbin and Sung, Jimin and Park, Sangah and Kwon, Heeyeon and Kwon, Taehee and Kim, Kanghyun and Kim, Namkug},
  title           = {Multimodal Large Language Models in Medical Imaging: Current State and Future Directions},
  journal         = {Korean Journal of Radiology},
  volume          = {26},
  number          = {10},
  pages           = {900--923},
  year            = {2025},
  month           = {10},
  publisher       = {The Korean Society of Radiology},
  issn            = {1229-6929},
  doi             = {10.3348/kjr.2025.0599},
  url             = {https://doi.org/10.3348/kjr.2025.0599}
}

@inproceedings{zhu2025can,
    title = "Can We Trust {AI} Doctors? A Survey of Medical Hallucination in Large Language and Large Vision-Language Models",
    author = "Zhu, Zhihong  and
      Zhang, Yunyan  and
      Zhuang, Xianwei  and
      Zhang, Fan  and
      Wan, Zhongwei  and
      Chen, Yuyan  and
      Long, Qingqing  and
      Zheng, Yefeng  and
      Wu, Xian",
    booktitle = "Findings of the Association for Computational Linguistics: ACL 2025",
    month = jul,
    year = "2025",
    address = "Vienna, Austria",
    publisher = "Association for Computational Linguistics",
    url = "https://aclanthology.org/2025.findings-acl.350/",
    doi = "10.18653/v1/2025.findings-acl.350",
    pages = "6748--6769",
    ISBN = "979-8-89176-256-5"
}

@article{bazi2023vision,
AUTHOR = {Bazi, Yakoub and Rahhal, Mohamad Mahmoud Al and Bashmal, Laila and Zuair, Mansour},
TITLE = {Vision–Language Model for Visual Question Answering in Medical Imagery},
JOURNAL = {Bioengineering},
VOLUME = {10},
YEAR = {2023},
NUMBER = {3},
ARTICLE-NUMBER = {380},
URL = {https://www.mdpi.com/2306-5354/10/3/380},
PubMedID = {36978771},
ISSN = {2306-5354},
DOI = {10.3390/bioengineering10030380}
}

@misc{kim2025medical,
      title={Medical Hallucinations in Foundation Models and Their Impact on Healthcare}, 
      author={Yubin Kim and Hyewon Jeong and Shan Chen and Shuyue Stella Li and Chanwoo Park and Mingyu Lu and Kumail Alhamoud and Jimin Mun and Cristina Grau and Minseok Jung and Rodrigo Gameiro and Chunjong Park and Hyeonhoon Lee and Hae Won Park and Daniel McDuff and Samir Tulebaev and Cynthia Breazeal},
      year={2025},
      eprint={2503.05777},
      archivePrefix={arXiv},
      primaryClass={cs.CL},
      url={https://arxiv.org/abs/2503.05777}, 
}

@article{miao2024chain,
AUTHOR = {Miao, Jing and Thongprayoon, Charat and Suppadungsuk, Supawadee and Krisanapan, Pajaree and Radhakrishnan, Yeshwanter and Cheungpasitporn, Wisit},
TITLE = {Chain of Thought Utilization in Large Language Models and Application in Nephrology},
JOURNAL = {Medicina},
VOLUME = {60},
YEAR = {2024},
NUMBER = {1},
ARTICLE-NUMBER = {148},
URL = {https://www.mdpi.com/1648-9144/60/1/148},
PubMedID = {38256408},
ISSN = {1648-9144},
DOI = {10.3390/medicina60010148}
}

@article{johnson2019mimic,
  title={MIMIC-CXR, a de-identified publicly available database of chest radiographs with free-text reports},
  author={Johnson, Alistair EW and Pollard, Tom J and Berkowitz, Seth J and Greenbaum, Nathaniel R and Lungren, Matthew P and Deng, Chih-ying and Mark, Roger G and Horng, Steven},
  journal={Scientific data},
  volume={6},
  number={1},
  pages={317},
  year={2019},
  publisher={Nature Publishing Group UK London}
}

@article{lau2018dataset,
  title={A dataset of clinically generated visual questions and answers about radiology images},
  author={Lau, Jason J and Gayen, Soumya and Ben Abacha, Asma and Demner-Fushman, Dina},
  journal={Scientific data},
  volume={5},
  number={1},
  pages={180251},
  year={2018},
  publisher={Nature Publishing Group}
}

@misc{med-hallmark,
      title={Detecting and Evaluating Medical Hallucinations in Large Vision Language Models}, 
      author={Jiawei Chen and Dingkang Yang and Tong Wu and Yue Jiang and Xiaolu Hou and Mingcheng Li and Shunli Wang and Dongling Xiao and Ke Li and Lihua Zhang},
      year={2024},
      eprint={2406.10185},
      archivePrefix={arXiv},
      primaryClass={cs.CV},
      url={https://arxiv.org/abs/2406.10185}, 
}

@inproceedings{eslami2023pubmedclip,
    title = "{P}ub{M}ed{CLIP}: How Much Does {CLIP} Benefit Visual Question Answering in the Medical Domain?",
    author = "Eslami, Sedigheh  and
      Meinel, Christoph  and
      de Melo, Gerard",
    booktitle = "Findings of the Association for Computational Linguistics: EACL 2023",
    month = may,
    year = "2023",
    address = "Dubrovnik, Croatia",
    publisher = "Association for Computational Linguistics",
    url = "https://aclanthology.org/2023.findings-eacl.88/",
    doi = "10.18653/v1/2023.findings-eacl.88",
    pages = "1181--1193"
}

@inproceedings{vaswani2017attention,
 author = {Vaswani, Ashish and Shazeer, Noam and Parmar, Niki and Uszkoreit, Jakob and Jones, Llion and Gomez, Aidan N and Kaiser, \L ukasz and Polosukhin, Illia},
 booktitle = {Advances in Neural Information Processing Systems},
 pages = {},
 publisher = {Curran Associates, Inc.},
 title = {Attention is All you Need},
 url = {https://proceedings.neurips.cc/paper_files/paper/2017/file/3f5ee243547dee91fbd053c1c4a845aa-Paper.pdf},
 volume = {30},
 year = {2017}
}

@article{ma2025debate,
title={Debate on Graph: A Flexible and Reliable Reasoning Framework for Large Language Models}, volume={39}, url={https://ojs.aaai.org/index.php/AAAI/article/view/34658}, DOI={10.1609/aaai.v39i23.34658},  number={23}, journal={Proceedings of the AAAI Conference on Artificial Intelligence}, author={Ma, Jie and Gao, Zhitao and Chai, Qi and Sun, Wangchun and Wang, Pinghui and Pei, Hongbin and Tao, Jing and Song, Lingyun and Liu, Jun and Zhang, Chen and Cui, Lizhen}, year={2025}, month={Apr.}, pages={24768-24776} }

@inproceedings{du2023improving,
author = {Du, Yilun and Li, Shuang and Torralba, Antonio and Tenenbaum, Joshua B. and Mordatch, Igor},
title = {Improving factuality and reasoning in language models through multiagent debate},
year = {2024},
publisher = {JMLR.org},
booktitle = {Proceedings of the 41st International Conference on Machine Learning},
articleno = {467},
numpages = {31},
location = {Vienna, Austria},
series = {ICML'24},url={https://openreview.net/pdf?id=zj7YuTE4t8}
}

@inproceedings{zhang2020counterfactual,
 author = {Zhang, Zhu and Zhao, Zhou and Lin, Zhijie and zhu, jieming and He, Xiuqiang},
 booktitle = {Advances in Neural Information Processing Systems},
 pages = {18123--18134},
 publisher = {Curran Associates, Inc.},
 title = {Counterfactual Contrastive Learning for Weakly-Supervised Vision-Language Grounding},
 url = {https://proceedings.neurips.cc/paper_files/paper/2020/file/d27b95cac4c27feb850aaa4070cc4675-Paper.pdf},
 volume = {33},
 year = {2020}
}

@inproceedings{zhong2022regionclip,
  author={Zhong, Yiwu and Yang, Jianwei and Zhang, Pengchuan and Li, Chunyuan and Codella, Noel and Li, Liunian Harold and Zhou, Luowei and Dai, Xiyang and Yuan, Lu and Li, Yin and Gao, Jianfeng},
  booktitle={2022 IEEE/CVF Conference on Computer Vision and Pattern Recognition (CVPR)}, 
  title={RegionCLIP: Region-based Language-Image Pretraining}, 
  year={2022},
  volume={},
  number={},
  pages={16772-16782},
  doi={10.1109/CVPR52688.2022.01629}}

@article{moor2023foundation,
  title={Foundation models for generalist medical artificial intelligence},
  author={Moor, Michael and Banerjee, Oishi and Abad, Zahra Shakeri Hossein and Krumholz, Harlan M and Leskovec, Jure and Topol, Eric J and Rajpurkar, Pranav},
  journal={Nature},
  volume={616},
  number={7956},
  pages={259--265},
  year={2023},
  publisher={Nature Publishing Group UK London},
  doi = {10.1038/s41586-023-05881-4},
  url ={https://doi.org/10.1038/s41586-023-05881-4}
}

@article{ji2023survey,
author = {Ji, Ziwei and Lee, Nayeon and Frieske, Rita and Yu, Tiezheng and Su, Dan and Xu, Yan and Ishii, Etsuko and Bang, Ye Jin and Madotto, Andrea and Fung, Pascale},
title = {Survey of Hallucination in Natural Language Generation},
year = {2023},
issue_date = {December 2023},
publisher = {Association for Computing Machinery},
address = {New York, NY, USA},
volume = {55},
number = {12},
issn = {0360-0300},
url = {https://doi.org/10.1145/3571730},
doi = {10.1145/3571730},
journal = {ACM Comput. Surv.},
month = mar,
articleno = {248},
numpages = {38}
}

@misc{liu2024survey,
      title={A Survey on Hallucination in Large Vision-Language Models}, 
      author={Hanchao Liu and Wenyuan Xue and Yifei Chen and Dapeng Chen and Xiutian Zhao and Ke Wang and Liping Hou and Rongjun Li and Wei Peng},
      year={2024},
      eprint={2402.00253},
      archivePrefix={arXiv},
      primaryClass={cs.CV},
      url={https://arxiv.org/abs/2402.00253}, 
}

@incollection{cohen2023potential,
  title={Patient safety in radiology and medical imaging},
  author={Ding, Alexander and Joshi, Jonathan and Tiwana, Emily},
  booktitle={Patient Safety: A Case-based Innovative Playbook for Safer Care},
  pages={261--277},
  year={2023},
  publisher={Springer}
}

@inproceedings{sun2024medhallmark,
    title = "{M}ed{H}allu: A Comprehensive Benchmark for Detecting Medical Hallucinations in Large Language Models",
    author = "Pandit, Shrey  and
      Xu, Jiawei  and
      Hong, Junyuan  and
      Wang, Zhangyang  and
      Chen, Tianlong  and
      Xu, Kaidi  and
      Ding, Ying",
    booktitle = "Proceedings of the 2025 Conference on Empirical Methods in Natural Language Processing",
    month = nov,
    year = "2025",
    address = "Suzhou, China",
    publisher = "Association for Computational Linguistics",
    url = "https://aclanthology.org/2025.emnlp-main.143/",
    doi = "10.18653/v1/2025.emnlp-main.143",
    pages = "2858--2873",
    ISBN = "979-8-89176-332-6"
}

@inproceedings{wei2022chain,
author = {Wei, Jason and Wang, Xuezhi and Schuurmans, Dale and Bosma, Maarten and Ichter, Brian and Xia, Fei and Chi, Ed H. and Le, Quoc V. and Zhou, Denny},
title = {Chain-of-thought prompting elicits reasoning in large language models},
year = {2022},
isbn = {9781713871088},
publisher = {Curran Associates Inc.},
address = {Red Hook, NY, USA},
booktitle = {Proceedings of the 36th International Conference on Neural Information Processing Systems},
articleno = {1800},
numpages = {14},
location = {New Orleans, LA, USA},
series = {NIPS '22},url={https://openreview.net/forum?id=_VjQlMeSB_J}
}

@article{singhal2023large,
  author          = {Singhal, Karan and Azizi, Shekoofeh and Tu, Tao and Mahdavi, S. Sara and Wei, Jason and Chung, Hyung Won and Scales, Nathan and Tanwani, Ajay and Cole-Lewis, Heather and Pfohl, Stephen and Payne, Perry and Seneviratne, Martin and Gamble, Paul and Kelly, Chris and Babiker, Alvin and Barral, Joelle and Semturs, Christopher and Karthikesalingam, Alan and Natarajan, Vivek},
  title           = {Large language models encode clinical knowledge},
  journal         = {Nature},
  year            = {2023},
  volume          = {620},
  number          = {7972},
  pages           = {172--180},
  month           = {08},
  doi             = {10.1038/s41586-023-06291-2},
  url             = {https://doi.org/10.1038/s41586-023-06291-2},
  issn            = {1476-4687}
}

@article{lievin2024can,
title = {Can large language models reason about medical questions?},
journal = {Patterns},
volume = {5},
number = {3},
pages = {100943},
year = {2024},
issn = {2666-3899},
doi = {https://doi.org/10.1016/j.patter.2024.100943},
url = {https://www.sciencedirect.com/science/article/pii/S2666389924000424},
author = {Valentin Liévin and Christoffer Egeberg Hother and Andreas Geert Motzfeldt and Ole Winther}
}

@article{liu2024lost,
    title = "Lost in the Middle: How Language Models Use Long Contexts",
    author = "Liu, Nelson F.  and
      Lin, Kevin  and
      Hewitt, John  and
      Paranjape, Ashwin  and
      Bevilacqua, Michele  and
      Petroni, Fabio  and
      Liang, Percy",
    journal = "Transactions of the Association for Computational Linguistics",
    volume = "12",
    year = "2024",
    address = "Cambridge, MA",
    publisher = "MIT Press",
    url = "https://aclanthology.org/2024.tacl-1.9/",
    doi = "10.1162/tacl_a_00638",
    pages = "157--173"
}

@inproceedings{wang2023self,
title={Self-Consistency Improves Chain of Thought Reasoning in Language Models},
author={Xuezhi Wang and Jason Wei and Dale Schuurmans and Quoc V Le and Ed H. Chi and Sharan Narang and Aakanksha Chowdhery and Denny Zhou},
booktitle={The Eleventh International Conference on Learning Representations },
year={2023},
url={https://openreview.net/forum?id=1PL1NIMMrw}
}

@inproceedings{liang2023encouraging,
    title = "Encouraging Divergent Thinking in Large Language Models through Multi-Agent Debate",
    author = "Liang, Tian  and
      He, Zhiwei  and
      Jiao, Wenxiang  and
      Wang, Xing  and
      Wang, Yan  and
      Wang, Rui  and
      Yang, Yujiu  and
      Shi, Shuming  and
      Tu, Zhaopeng",
    booktitle = "Proceedings of the 2024 Conference on Empirical Methods in Natural Language Processing",
    month = nov,
    year = "2024",
    address = "Miami, Florida, USA",
    publisher = "Association for Computational Linguistics",
    url = "https://aclanthology.org/2024.emnlp-main.992/",
    doi = "10.18653/v1/2024.emnlp-main.992",
    pages = "17889--17904"
}

@inproceedings{li2023camel,
author = {Li, Guohao and Al Kader Hammoud, Hasan Abed and Itani, Hani and Khizbullin, Dmitrii and Ghanem, Bernard},
title = {CAMEL: communicative agents for "mind" exploration of large language model society},
year = {2023},
publisher = {Curran Associates Inc.},
address = {Red Hook, NY, USA},
booktitle = {Proceedings of the 37th International Conference on Neural Information Processing Systems},
articleno = {2264},
numpages = {18},
location = {New Orleans, LA, USA},
series = {NIPS '23},url={https://openreview.net/forum?id=3IyL2XWDkG}
}

@inproceedings{tang2024medagents,
    title = "{M}ed{A}gents: Large Language Models as Collaborators for Zero-shot Medical Reasoning",
    author = "Tang, Xiangru  and
      Zou, Anni  and
      Zhang, Zhuosheng  and
      Li, Ziming  and
      Zhao, Yilun  and
      Zhang, Xingyao  and
      Cohan, Arman  and
      Gerstein, Mark",
    booktitle = "Findings of the Association for Computational Linguistics: ACL 2024",
    month = aug,
    year = "2024",
    address = "Bangkok, Thailand",
    publisher = "Association for Computational Linguistics",
    url = "https://aclanthology.org/2024.findings-acl.33/",
    doi = "10.18653/v1/2024.findings-acl.33",
    pages = "599--621"
}

@book{popper2005logic,
  title={The logic of scientific discovery},
  author={Popper, Karl},
  year={2005},
  publisher={Routledge}
}

@inproceedings{niu2021counterfactual,
  author={Niu, Yulei and Tang, Kaihua and Zhang, Hanwang and Lu, Zhiwu and Hua, Xian-Sheng and Wen, Ji-Rong},
  booktitle={2021 IEEE/CVF Conference on Computer Vision and Pattern Recognition (CVPR)}, 
  title={Counterfactual VQA: A Cause-Effect Look at Language Bias}, 
  year={2021},
  volume={},
  number={},
  pages={12695-12705},
  doi={10.1109/CVPR46437.2021.01251}}

@inproceedings{boecking2022making,
author="Boecking, Benedikt
and Usuyama, Naoto
and Bannur, Shruthi
and Castro, Daniel C.
and Schwaighofer, Anton
and Hyland, Stephanie
and Wetscherek, Maria
and Naumann, Tristan
and Nori, Aditya
and Alvarez-Valle, Javier
and Poon, Hoifung
and Oktay, Ozan",
editor="Avidan, Shai
and Brostow, Gabriel
and Ciss{\'e}, Moustapha
and Farinella, Giovanni Maria
and Hassner, Tal",
title = {Making the most of text semantics to improve biomedical vision--language processing},
year = {2022},
isbn = {978-3-031-20058-8},
publisher = {Springer-Verlag},
address = {Berlin, Heidelberg},
url = {https://doi.org/10.1007/978-3-031-20059-5_1},
doi = {10.1007/978-3-031-20059-5_1},
booktitle = {Computer Vision – ECCV 2022: 17th European Conference, Tel Aviv, Israel, October 23–27, 2022, Proceedings, Part XXXVI},
pages = {1–21},
numpages = {21},
location = {Tel Aviv, Israel}
}

@article{nickerson1998confirmation,
  title={Confirmation bias: A ubiquitous phenomenon in many guises},
  author={Nickerson, Raymond S},
  journal={Review of general psychology},
  volume={2},
  number={2},
  pages={175--220},
  year={1998},
  publisher={SAGE Publications Sage CA: Los Angeles, CA}
}

@misc{he2020pathvqa,
      title={PathVQA: 30000+ Questions for Medical Visual Question Answering}, 
      author={Xuehai He and Yichen Zhang and Luntian Mou and Eric Xing and Pengtao Xie},
      year={2020},
      eprint={2003.10286},
      archivePrefix={arXiv},
      primaryClass={cs.CL},
      url={https://arxiv.org/abs/2003.10286}, 
}

@inproceedings{li2024llava,
author = {Li, Chunyuan and Wong, Cliff and Zhang, Sheng and Usuyama, Naoto and Liu, Haotian and Yang, Jianwei and Naumann, Tristan and Poon, Hoifung and Gao, Jianfeng},
title = {LLaVA-med: training a large language-and-vision assistant for biomedicine in one day},
year = {2023},
publisher = {Curran Associates Inc.},
address = {Red Hook, NY, USA},
url={https://openreview.net/forum?id=GSuP99u2kR},
booktitle = {Proceedings of the 37th International Conference on Neural Information Processing Systems},
articleno = {1240},
numpages = {24},
location = {New Orleans, LA, USA},
series = {NIPS '23}
}

@inproceedings{chen2024reconcile,
    title = "{R}e{C}oncile: Round-Table Conference Improves Reasoning via Consensus among Diverse {LLM}s",
    author = "Chen, Justin  and
      Saha, Swarnadeep  and
      Bansal, Mohit",
    booktitle = "Proceedings of the 62nd Annual Meeting of the Association for Computational Linguistics (Volume 1: Long Papers)",
    month = aug,
    year = "2024",
    address = "Bangkok, Thailand",
    publisher = "Association for Computational Linguistics",
    url = "https://aclanthology.org/2024.acl-long.381/",
    doi = "10.18653/v1/2024.acl-long.381",
    pages = "7066--7085"
}

@article{pelka2018roco,
   title={ROCOv2: Radiology Objects in COntext Version 2, an Updated Multimodal Image Dataset},
   volume={11},
   ISSN={2052-4463},
   url={http://dx.doi.org/10.1038/s41597-024-03496-6},
   DOI={10.1038/s41597-024-03496-6},
   number={1},
   journal={Scientific Data},
   publisher={Springer Science and Business Media LLC},
   author={Rückert, Johannes and Bloch, Louise and Brüngel, Raphael and Idrissi-Yaghir, Ahmad and Schäfer, Henning and Schmidt, Cynthia S. and Koitka, Sven and Pelka, Obioma and Abacha, Asma Ben and G. Seco de Herrera, Alba and Müller, Henning and Horn, Peter A. and Nensa, Felix and Friedrich, Christoph M.},
   year={2024},
   month=jun }

@inproceedings{rohrbach2018object,
    title = "Object Hallucination in Image Captioning",
    author = "Rohrbach, Anna  and
      Hendricks, Lisa Anne  and
      Burns, Kaylee  and
      Darrell, Trevor  and
      Saenko, Kate",
    booktitle = "Proceedings of the 2018 Conference on Empirical Methods in Natural Language Processing",
    month = oct # "-" # nov,
    year = "2018",
    address = "Brussels, Belgium",
    publisher = "Association for Computational Linguistics",
    url = "https://aclanthology.org/D18-1437/",
    doi = "10.18653/v1/D18-1437",
    pages = "4035--4045"
}

@inproceedings{bae2023ehrxqamultimodalquestionanswering,
 author = {Bae, Seongsu and Kyung, Daeun and Ryu, Jaehee and Cho, Eunbyeol and Lee, Gyubok and Kweon, Sunjun and Oh, Jungwoo and Ji, Lei and Chang, Eric and Kim, Tackeun and Choi, Edward},
 booktitle = {Advances in Neural Information Processing Systems},
 pages = {3867--3880},
 publisher = {Curran Associates, Inc.},
 title = {EHRXQA: A Multi-Modal Question Answering Dataset for Electronic Health Records with Chest X-ray Images},
 url = {https://proceedings.neurips.cc/paper_files/paper/2023/file/0c007ebef1d11fd48da6ce4f54687db6-Paper-Datasets_and_Benchmarks.pdf},
 volume = {36},
 year = {2023}
}

@article{medvcd,
   title={Med-VCD: Mitigating hallucination for medical large vision language models through visual contrastive decoding},
   volume={200},
   ISSN={0010-4825},
   url={http://dx.doi.org/10.1016/j.compbiomed.2025.111347},
   DOI={10.1016/j.compbiomed.2025.111347},
   journal={Computers in Biology and Medicine},
   publisher={Elsevier BV},
   author={Mahdavi, Zahra and Khodakaramimaghsoud, Zahra and Khaloo, Hooman and Taleshani, Sina Bakhshandeh and Hashemi, Erfan and Kaleybar, Javad Mirzapour and Manzari, Omid Nejati},
   year={2026},
   month=jan, pages={111347} }

@article{deepgbtb2026, title={DeepGB-TB: A Risk-Balanced Cross-Attention Gradient-Boosted Convolutional Network for Rapid, Interpretable Tuberculosis Screening}, volume={40}, url={https://ojs.aaai.org/index.php/AAAI/article/view/41245}, DOI={10.1609/aaai.v40i46.41245}, number={46}, journal={Proceedings of the AAAI Conference on Artificial Intelligence}, author={Lu, Zhixiang and Li, Yulong and Tang, Feilong and Jiang, Zhengyong and Li, Chong and Zhou, Mian and Li, Tenglong and Su, Jionglong}, year={2026}, month={Mar.}, pages={38989-38997} }

@misc{tang2026causalsamllmlargelanguagemodels,
      title={Causal-SAM-LLM: Large Language Models as Causal Reasoners for Robust Medical Segmentation}, 
      author={Tao Tang and Shijie Xu and Jionglong Su and Zhixiang Lu},
      year={2026},
      eprint={2507.03585},
      archivePrefix={arXiv},
      primaryClass={cs.CV},
      url={https://arxiv.org/abs/2507.03585}, 
}

@misc{lu2026skinclipvlconsistencyawarevisionlanguagelearning,
      title={SkinCLIP-VL: Consistency-Aware Vision-Language Learning for Multimodal Skin Cancer Diagnosis}, 
      author={Zhixiang Lu and Shijie Xu and Kaicheng Yan and Xuyue Cai and Chong Zhang and Yulong Li and Angelos Stefanidis and Anh Nguyen and Jionglong Su},
      year={2026},
      eprint={2603.21010},
      archivePrefix={arXiv},
      primaryClass={cs.CV},
      url={https://arxiv.org/abs/2603.21010}, 
}
\appendix

\section{Algorithm Summary}
\label{algorithmsum}
The overall process is summarized in Algorithm~\ref{alg:dialectic_med}.

\subsection{Dynamic Consensus Graph Formulation}
\label{sec:graph_formulation}

To explicitly model the dialectic trajectory and mitigate the ``lost-in-the-middle'' phenomenon common in long-context reasoning \cite{liu2024lost}, we introduce the \textbf{Dynamic Consensus Graph}, denoted as $\mathcal{G}_t = (\mathcal{V}_t, \mathcal{E}_t)$ at time step $t$. This Directed Acyclic Graph (DAG) serves as a structured memory bank that records the evolution of diagnostic hypotheses under adversarial scrutiny, enabling principled consensus aggregation \cite{ma2025debate}.

\subsubsection{Graph Definition}
The node set $\mathcal{V}_t$ consists of two distinct types of logical entities:
\begin{itemize}
    \item \textbf{Hypothesis nodes} $h \in \mathcal{H}$: Diagnostic hypotheses generated by the Proponent agent $\mathcal{A}_P$, each associated with a semantic embedding $\mathbf{h} \in \mathbb{R}^d$.
    \item \textbf{Counter-Evidence nodes} $e \in \mathcal{E}_{\text{vid}}$: Visually-grounded counter-arguments derived by the Opponent agent $\mathcal{A}_O$ via the Visual Falsification Module.
\end{itemize}

The edge set $\mathcal{E}_t$ represents the logical transitions between nodes. We define two types of directed edges:
\begin{itemize}
    \item \textbf{Falsification edges} $(h_i, e_j)$: Connecting a hypothesis to the counter-evidence that challenges it.
    \item \textbf{Rectification edges} $(e_j, h_k)$: Connecting counter-evidence to the revised hypothesis it prompted.
\end{itemize}

Associated with each edge $(u, v) \in \mathcal{E}_t$ is a transition weight $w_{uv} \in [0, 1]$, quantifying the confidence of the logical inference. At $t=0$, the graph is initialized with the primary hypothesis $h_0$:
\begin{equation}
    \mathcal{V}_0 = \{h_0\}, \quad \mathcal{E}_0 = \emptyset
\end{equation}

\subsubsection{Adversarial Graph Expansion}
In each dialectic iteration $t$, given the current hypothesis $h_{t-1}$, the Opponent $\mathcal{A}_O$ mines a visual counter-evidence $e_t$ using the Visual Falsification Module. The Mediator $\mathcal{A}_M$ evaluates the validity of this attack by computing the \textbf{Attack Strength} $S_{\text{attack}}$, which measures the visual grounding confidence of the counter-argument:
\begin{equation}
    S_{\text{attack}}(e_t) = \frac{1}{|R(e_t)|} \sum_{k \in R(e_t)} \alpha_k^{(t)}
\end{equation}
where $R(e_t)$ denotes the set of image patch indices referenced by the counter-evidence $e_t$, and $\alpha_k^{(t)}$ is the attention weight from the falsification attention map $M_F^{(t)}$.

If $S_{\text{attack}}(e_t) \geq \theta_{\text{attack}}$ (i.e., the attack is deemed valid), the Proponent generates a revised hypothesis $h_t$, and the graph is expanded:
\begin{equation}
    \mathcal{V}_t = \mathcal{V}_{t-1} \cup \{ e_t, h_t \}
\end{equation}
\begin{equation}
    \mathcal{E}_t = \mathcal{E}_{t-1} \cup \{ (h_{t-1}, e_t), (e_t, h_t) \}
\end{equation}

The edge weights are assigned as follows:
\begin{itemize}
    \item The \textit{falsification edge} $(h_{t-1}, e_t)$ is weighted by the attack strength: $w_{h_{t-1}, e_t} = S_{\text{attack}}(e_t)$.
    \item The \textit{rectification edge} $(e_t, h_t)$ is weighted by the Proponent's revised confidence: $w_{e_t, h_t} = \text{conf}(h_t | e_t)$, where $\text{conf}(\cdot)$ is derived from the Proponent's output logits.
\end{itemize}

\subsubsection{Path Integration}
To derive the final diagnosis $D_{\text{final}}$, we perform a probabilistic aggregation over all paths in the final graph $\mathcal{G}_T$. Let $\Pi(h_{\text{leaf}})$ denote the set of all directed paths from the root hypothesis $h_0$ to a leaf hypothesis node $h_{\text{leaf}} \in \mathcal{H}_{\text{leaf}}$. The cumulative \textbf{Credibility Score} $\Phi$ for a terminal hypothesis is calculated by integrating the transition weights along each reasoning chain:
\begin{equation}
\small
\Phi(h_{leaf}) = \sum_{\pi \in \Pi(h_{leaf})} \exp \left( \frac{1}{|\pi|} \sum_{(u,v) \in \pi} \log(w_{uv}) \right)
\end{equation}
This formulation ensures that a diagnosis is only reliable if it survives the adversarial loop with high-confidence transitions at every step. Intuitively, a hypothesis that was revised due to strongly grounded counter-evidence (high $w_{h, e}$) and then confidently restated (high $w_{e, h'}$) will accumulate a high credibility score.

The final diagnosis is selected via:
\begin{equation}
    D_{\text{final}} = \operatorname*{arg\,max}_{h \in \mathcal{H}_{\text{leaf}}} \Phi(h)
\label{eq:final_diag}
\end{equation}

The confidence of the final diagnosis is normalized:
\begin{equation}
    \text{Confidence}(D_{\text{final}}) = \frac{\Phi(D_{\text{final}})}{\sum_{h \in \mathcal{H}_{\text{leaf}}} \Phi(h)}
\end{equation}

\paragraph{Cycle Detection and Pruning.} To maintain the DAG property and prevent infinite loops, if the dialectic process proposes a hypothesis $h_t$ that is semantically equivalent to a previously refuted hypothesis $h_j$ (i.e., $\text{sim}(h_t, h_j) > \theta_{\text{sim}}$), the branch is pruned. This is implemented by checking against all existing hypothesis nodes before adding $h_t$ to $\mathcal{V}_t$.

\subsubsection{Explanation Generation}
The final explanation $E$ is constructed by tracing the highest-scoring path in $\mathcal{G}_T$ that leads to $D_{\text{final}}$. The Mediator $\mathcal{A}_M$ summarizes this path, explicitly referencing the key counter-evidence nodes and the visual regions they highlighted. This ensures the explanation is both comprehensive and verifiable, providing a transparent audit trail of the diagnostic reasoning process.

\subsection{Inference Algorithm}
\label{sec:inference_algorithm}

The complete inference process of our Dialectic-Med framework is outlined in Algorithm~\ref{alg:dialectic_med}. The algorithm orchestrates the three agents through the adversarial dialectic loop, dynamically updates the consensus graph, and finally aggregates the results to produce a robust diagnosis.

\begin{algorithm}[t!]
\caption{Adversarial Dialectic Reasoning with Consensus Graph}
\label{alg:dialectic_med}
\KwData{Medical Image $I$, User Query $Q$, Max Iterations $T_{\max}$, Attack Threshold $\theta_{\text{attack}}$, Similarity Threshold $\theta_{\text{sim}}$}
\KwResult{Final Diagnosis $D_{\text{final}}$, Explanation $E$, Confidence $C$}

\textbf{Initialize:} Agents $\mathcal{A}_P, \mathcal{A}_O, \mathcal{A}_M$\;
$h_0 \gets \mathcal{A}_P.\text{Generate}(I, Q)$; 
$\mathcal{G} \gets \text{InitGraph}(h_0)$; $\mathcal{V} \gets \{h_0\}$; $\mathcal{E} \gets \emptyset$\;
$h_{\text{current}} \gets h_0$; $t \gets 1$\;

\While{$t \leq T_{\max}$}{
  $Q_O^{(t)} \gets \mathcal{A}_O.\text{GenerateProbe}(h_{\text{current}})$\;
  $M_F^{(t)} \gets \text{VFM}(I, Q_O^{(t)})$;
  $e_t \gets \mathcal{A}_O.\text{Generate}(h_{\text{current}}, M_F^{(t)})$\;

  $S_{\text{attack}} \gets \text{Attack}(e_t, M_F^{(t)})$;

  \lIf{$S_{\text{attack}} < \theta_{\text{attack}}$}{
    \textbf{break};
  }

  $h_t \gets \mathcal{A}_P.\text{Revise}(h_{\text{current}}, e_t)$\;

  \lIf{$\exists h_j \in \mathcal{V}: \text{sim}(h_t, h_j) > \theta_{\text{sim}}$}{
    \textbf{continue}; 
  }

  $\mathcal{V} \gets \mathcal{V} \cup \{e_t, h_t\}$\;
  $\mathcal{E} \gets \mathcal{E} \cup \{(h_{\text{current}}, e_t), (e_t, h_t)\}$\;
  $w_{h_{\text{current}}, e_t} \gets S_{\text{attack}}$\;
  $w_{e_t, h_t} \gets \mathcal{A}_P.\text{Confidence}(h_t)$\;

  $h_{\text{current}} \gets h_t$; $t \gets t + 1$\;
}

$\mathcal{H}_{\text{leaf}} \gets \text{GetLeafHypotheses}(\mathcal{G})$\;
\ForEach{$h \in \mathcal{H}_{\text{leaf}}$}{
  $\Phi(h) \gets \text{Credibility}(h, \mathcal{G})$;
}
$D_{\text{final}} \gets \operatorname*{arg\,max}_{h \in \mathcal{H}_{\text{leaf}}} \Phi(h)$;
$C \gets \Phi(D_{\text{final}}) / \sum_{h} \Phi(h)$\;
$E \gets \mathcal{A}_M.\text{SummarizePath}(\mathcal{G}, D_{\text{final}})$\;

\Return $D_{\text{final}}, E, C$\;
\end{algorithm}

\paragraph{Complexity Analysis.} The time complexity of Algorithm~\ref{alg:dialectic_med} is $O(T_{\max} \cdot (C_{\text{VFM}} + C_{\text{LLM}}))$, where $C_{\text{VFM}}$ is the cost of a single VFM forward pass and $C_{\text{LLM}}$ is the cost of an LLM inference. The graph operations (adding nodes, edges, and computing credibility) are $O(|\mathcal{V}|^2)$ in the worst case, but since $|\mathcal{V}| \leq 2T_{\max} + 1$, this is dominated by the LLM inference cost. In practice, we set $T_{\max} \in [3, 5]$, making the overhead minimal compared to single-pass MLLM inference while significantly improving diagnostic reliability.

\subsection{Visual Falsification Module (VFM)}
The Visual Falsification Module is the cornerstone of the Opponent agent's ability to challenge the Proponent. It is designed to operationalize the principle of falsification by actively seeking and localizing visual evidence that contradicts a given diagnostic hypothesis. The VFM comprises two main components: a Counterfactual Probe Generator and a Grounded Attention Mechanism, which are fine-tuned using a novel counterfactual grounding objective.

\subsubsection{Network Architecture}
The VFM leverages a pre-trained VLM with a Vision Transformer (ViT) backbone (PubMedCLIP \cite{eslami2023pubmedclip}), which is specifically adapted for the medical domain. The architecture includes:
\begin{itemize}
    \item \textbf{Visual Encoder ($\mathcal{E}_v$):} A ViT model that processes an input image $I$ by dividing it into a sequence of $N$ flattened 2D patches, $\{p_1, p_2, ..., p_N\}$. Each patch is linearly projected into a patch embedding. Including the [CLS] token, the output is a sequence of patch embeddings $\mathbf{V} = \{v_{\text{cls}}, v_1, ..., v_N\} \in \mathbb{R}^{(N+1) \times D}$, where $D$ is the embedding dimension.
    \item \textbf{Textual Encoder ($\mathcal{E}_t$):} A Transformer-based text encoder that processes the counterfactual probe query $Q_O$ and outputs a sentence embedding $\mathbf{q} \in \mathbb{R}^D$.
    \item \textbf{Cross-Modal Attention Layer:} A standard cross-modal attention mechanism \cite{deepgbtb2026} that computes the similarity between the textual probe embedding and each of the visual patch embeddings.
\end{itemize}
This dual-encoder architecture allows us to project both the image regions and the textual query into a shared embedding space, enabling fine-grained, region-level semantic matching \cite{zhong2022regionclip}.

\subsubsection{Mathematical Formulation}
The core function of the VFM is to generate a \textbf{Falsification Attention Map} ($M_F$) that highlights image regions inconsistent with the Proponent's hypothesis $H_P$. This is achieved through a two-step process:

\paragraph{Counterfactual Probe Generation.} Given $H_P$, the Opponent agent $\mathcal{A}_O$ first generates a textual counterfactual probe $Q_O$. This is not a simple negation, but a targeted query for contradictory evidence, leveraging medical domain knowledge $\mathcal{K}$. For a hypothesis $H_P =$ ``Left lower lobe opacity consistent with Pneumonia'', the generation process can be modeled as:

\begin{equation}
    Q_O = \mathcal{A}_O(\mathcal{T}_{\text{probe}}(H_P, \mathcal{K}))
    \label{eq:opponent_query}
\end{equation}
This yields a probe like: ``\textit{Identify features contradicting pneumonia, specifically signs of volume loss}".

\paragraph{Grounded Attention via Cross-Modal Similarity.} The VFM then grounds this probe $Q_O$ in the image $I$. We compute the cross-modal similarity between the probe's text embedding $\mathbf{q} = \mathcal{E}_t(Q_O)$ and each visual patch embedding $v_i \in \mathbf{V}$. The relevance score $s_i$ for each patch $i$ is calculated using the scaled dot-product attention mechanism \cite{vaswani2017attention}:
\begin{equation}
    s_i = \frac{\mathbf{q}^T v_i}{\|\mathbf{q}\| \|v_i\| \sqrt{d}}
\end{equation}

where the cosine similarity is scaled by the square root of the embedding dimension $d$ to stabilize gradients. These scores represent the semantic alignment between the counterfactual probe and each image region.

The raw scores are then normalized using a softmax function to produce the final Falsification Attention Map $M_F = \{\alpha_1, \alpha_2, ..., \alpha_N\}$:
\begin{equation}
    \alpha_i = \frac{\exp(s_i / \tau)}{\sum_{j=1}^{N} \exp(s_j / \tau)}
\label{eq:falsification_attention}
\end{equation}
where $\tau$ is a temperature parameter. A lower $\tau$ produces a sharper attention map, focusing on the most salient contradictory evidence. This process is analogous to generating a saliency map, but instead of highlighting regions that \textit{support} a classification, it highlights regions that \textit{falsify} it, similar to techniques used in counterfactual visual explanations.

\paragraph{Generating Counter-Evidence.} The Opponent agent $\mathcal{A}_O$ uses this attention map to generate its counter-argument $Arg_O$. It focuses on the top-$k$ patches with the highest attention weights, denoted as $R_k = \{\text{patch}_i | \alpha_i \in \text{TopK}(\alpha)\}$. The final counter-argument is generated by prompting the agent with the original hypothesis and the localized visual evidence:
\begin{equation}
    Arg_O = \mathcal{A}_O(\mathcal{T}_{\text{refute}}(H_P, R_k))
    \label{eq:opponent_arg}
\end{equation}
This ensures that the Opponent's argument is not a generic rebuttal but is directly and verifiably grounded in specific, localized visual features, as shown in the case study in Figure~\ref{fig:case} where the falsification attention correctly localizes the elevated diaphragm.

\subsubsection{Training Objective}
While the VFM can operate in a zero-shot manner using a pre-trained VLM, its ability to localize subtle, contradictory findings can be significantly enhanced through fine-tuning. We propose a novel \textbf{Counterfactual Grounding Loss} ($\mathcal{L}_{\text{CFG}}$) designed to train the VFM to explicitly distinguish between visual evidence that supports a hypothesis and evidence that falsifies it.

To construct training triplets, we require a dataset with bounding box annotations for both positive and negative findings. Given an image $I$, a diagnostic hypothesis $H_P$, a set of ground-truth bounding boxes $B_P$ for findings consistent with $H_P$, and a set of bounding boxes $B_O$ for findings that contradict $H_P$, we can formulate our training objective.

Let $M_P$ be the attention map generated by a standard probe for $H_P$ (e.g., ``Find signs of pneumonia"), and let $M_F$ be the falsification attention map generated by the counterfactual probe $Q_O$. Our goal is to train the model such that:
\begin{itemize}
    \item The attention map $M_P$ focuses on the regions defined by $B_P$.
    \item The falsification map $M_F$ focuses on the regions defined by $B_O$.
\end{itemize}

We adapt the contrastive learning framework from \cite{zhang2020counterfactual} for this purpose. For a given hypothesis $H_P$, we define:
\begin{itemize}
    \item \textbf{Positive Alignment Score ($S^+$):} The aggregated attention from the standard probe map $M_P$ within the positive bounding boxes $B_P$.
    \item \textbf{Negative Alignment Score ($S^-$):} The aggregated attention from the standard probe map $M_P$ within the \textit{negative} (contradictory) bounding boxes $B_O$.
\end{itemize}

Our first loss term, the \textbf{Proponent Grounding Loss} ($\mathcal{L}_{P}$), encourages the standard attention map to focus on the correct evidence:
\begin{equation}
    \mathcal{L}_{P} = -\log \frac{\text{exp}(S^+ / \tau)}{\text{exp}(S^+ / \tau) + \text{exp}(S^- / \tau)}
\end{equation}

Symmetrically, for the falsification map $M_F$ generated by the counterfactual probe $Q_O$, we define:
\begin{itemize}
    \item \textbf{Falsification Alignment Score ($S_F^+$):} The aggregated attention from the falsification map $M_F$ within the contradictory bounding boxes $B_O$.
    \item \textbf{Falsification Misalignment Score ($S_F^-$):} The aggregated attention from the falsification map $M_F$ within the \textit{positive} bounding boxes $B_P$.
\end{itemize}

Our second loss term, the \textbf{Opponent Grounding Loss} ($\mathcal{L}_{O}$), trains the VFM to correctly localize the counter-evidence:
\begin{equation}
    \mathcal{L}_{O} = -\log \frac{\text{exp}(S_F^+ / \tau)}{\text{exp}(S_F^+ / \tau) + \text{exp}(S_F^- / \tau)}
\end{equation}

The total Counterfactual Grounding Loss is the sum of these two components:
\begin{equation}
    \mathcal{L}_{\text{CFG}} = \lambda_P \mathcal{L}_{P} + \lambda_O \mathcal{L}_{O}
\label{eq:cfg_loss}
\end{equation}
where $\lambda_P$ and $\lambda_O$ are hyperparameters to balance the two objectives. By minimizing this loss, we explicitly train the VFM to not only identify supporting evidence but also to become an expert at seeking out and localizing contradictory visual signals. This dual objective is critical for breaking the cycle of confirmation bias that plagues standard MLLMs.

The overall training objective for the entire Dialectic-Med framework combines this with the standard language modeling loss:
\begin{equation}
    \mathcal{L}_{\text{total}} = \mathcal{L}_{\text{LM}} + \beta \mathcal{L}_{\text{CFG}}
\end{equation}
where $\beta$ is a balancing hyperparameter.

\section{Details of Expert Human Evaluation}
\label{app:human_eval}

To rigorously assess the clinical safety and explanation faithfulness of our framework, we conducted a double-blind human evaluation as reported in Section \ref{quantitative}. To comply with responsible research guidelines, we provide the following logistical details:

\paragraph{Recruitment and Compensation.} The evaluation was conducted by three board-certified clinical radiologists with an average of 8 years of clinical experience. They were recruited through existing academic and clinical collaboration networks. Participants were compensated for their time at standard institutional clinical consulting rates, which is commensurate with their professional expertise and demographic location.

\paragraph{Evaluation Protocol and Instructions.} The task did not involve human subjects research, as evaluators only assessed AI-generated textual outputs based on de-identified public images. Evaluators were presented with the input X-ray, the ground-truth report findings, and the reasoning trajectories generated by different anonymized models (Dialectic-Med, MedVCD, and standard GPT-4o CoT). They were instructed to score the \textit{Explanation Faithfulness} on a 1-5 scale based on the following rubric:
\begin{itemize}[leftmargin=*, nosep]
    \item \textbf{1 (Dangerous):} The explanation contains severe, visually ungrounded hallucinations that could lead to clinical harm.
    \item \textbf{3 (Mixed):} The reasoning is generally correct but includes minor fabricated visual details.
    \item \textbf{5 (Highly Faithful):} The reasoning is strictly grounded in verified visual evidence, accurately addressing counterfactuals without any hallucinations.
\end{itemize}

\onecolumn
\section{Prompt Example of Agents}
\label{sec:prompts}
\begin{agentbox}{propFrame}{Proponent Agent ($\mathcal{A}_P$)}
    \textbf{Role:} Similar to a primary care physician, focusing on the \textbf{Global Context} of the image.\\
    \textbf{Task:} Propose an initial hypothesis or revise the hypothesis when facing counter-evidence.
    
    \noindent\rule{\textwidth}{0.5pt} 
    
    \textbf{SYSTEM PROMPT} \\
    \texttt{You are an experienced Radiologist acting as the "Proponent Agent". Your goal is to provide the most probable diagnosis based on the medical image analysis.}
    
    \textbf{Guidelines:}
    \begin{itemize}[noitemsep, topsep=0pt]
        \item \textbf{Focus on Global Context:} Look at the overall opacity, lung volume, and heart size.
        \item \textbf{Be Open-Minded:} You will receive counter-arguments from an Opponent. If their visual evidence is strong, acknowledge it and revise your hypothesis.
        \item \textbf{Logical Reasoning:} Explain your reasoning step-by-step before giving the final diagnosis label.
    \end{itemize}
    
    \vspace{0.2cm}
    \textbf{USER PROMPT (Iteration $t=0$: Initialization)} \\
    \texttt{<Image Context>} \\
    \texttt{Global Visual Features: \{\{GLOBAL\_FEATURES\_DESCRIPTION\}\}} \\
    \texttt{User Query: \{\{USER\_QUERY\}\}} \\
    \texttt{</Image Context>}
    
    Based on the global visual features, provide an initial hypothesis ($H_0$). \\
    \textbf{Output Format:}
    \begin{itemize}[noitemsep]
        \item Reasoning: [Your analysis]
        \item Hypothesis: [Diagnosis Name]
        \item Confidence: [0-100\%]
    \end{itemize}

    \vspace{0.2cm}
    \textbf{USER PROMPT (Iteration $t>0$: Revision)} \\
    \texttt{Current Hypothesis ($H_{t-1}$): \{\{CURRENT\_HYPOTHESIS\}\}} \\
    \texttt{Opponent's Counter-Argument: "\{\{OPPONENT\_ARGUMENT\}\}"} \\
    \texttt{Local Visual Evidence: \{\{LOCAL\_VISUAL\_FEATURES\}\}}
    
    \textbf{Instruction:} The Opponent claims the local evidence contradicts your hypothesis.
    \begin{enumerate}[noitemsep]
        \item Evaluate if the counter-argument is valid.
        \item If valid, propose a Revised Hypothesis ($H_t$) that explains both global context and local detail.
        \item If invalid, defend your original hypothesis.
    \end{enumerate}
\end{agentbox}

\newpage

\begin{agentbox}{oppFrame}{Opponent Agent ($\mathcal{A}_O$)}
    \textbf{Role:} Similar to a "Medical Auditor", focusing on \textbf{Visual Falsification}.\\
    \textbf{Task:} Use a "Visual Probe" to find local features that contradict the current hypothesis.
    
    \noindent\rule{\textwidth}{0.5pt}
    
    \textbf{SYSTEM PROMPT} \\
    \texttt{You are a critical Medical Auditor acting as the "Opponent Agent". Your ONLY goal is to \textbf{FALSIFY} the current diagnosis hypothesis. You utilize a "Visual Falsification Module" to probe specific regions.}
    
    \textbf{Guidelines:}
    \begin{itemize}[noitemsep, topsep=0pt]
        \item \textbf{Seek Contradictions:} Do not look for confirming evidence. Look for what is WRONG with the hypothesis.
        \item \textbf{Focus on Local Detail:} Use the provided local visual probe data (e.g., costophrenic angles).
        \item \textbf{Be Sharp:} Your argument must be grounded in the visual evidence provided.
    \end{itemize}
    
    \vspace{0.2cm}
    \textbf{USER PROMPT} \\
    \texttt{Current Hypothesis ($H_t$): \{\{CURRENT\_HYPOTHESIS\}\}} \\
    \texttt{Visual Probe Target: ROI focused on \{\{ROI\_NAME\}\}.} \\
    \texttt{Local Visual Features: \{\{LOCAL\_FEATURES\_DESCRIPTION\}\}}
    
    \textbf{Instruction:} Does the visual evidence in this ROI contradict the Current Hypothesis?
    \begin{itemize}[noitemsep]
        \item If hypothesis is "Pneumonia" (implies opacity), but ROI shows "Sharp Costophrenic Angle", this is a contradiction.
        \item If hypothesis is "Normal", but ROI shows "Nodule", this is a contradiction.
    \end{itemize}
    
    Generate a Counter-Argument ($Arg_{opp}$). \\
    \textbf{Output Format:}
    \begin{itemize}[noitemsep]
        \item Observation: "I see [Visual Feature] in the [Region]..."
        \item Contradiction: "This contradicts [Hypothesis] because [Reason]..."
        \item Counter-Evidence Strength: [High/Medium/Low]
    \end{itemize}
\end{agentbox}

\newpage

\begin{agentbox}{medFrame}{Mediator Agent ($\mathcal{A}_M$)}
    \textbf{Role:} Similar to a Senior Consultant or Judge, responsible for \textbf{Consensus Aggregation}.\\
    \textbf{Task:} Evaluate the debate quality, decide on revisions, and determine consensus.
    
    \noindent\rule{\textwidth}{0.5pt}
    
    \textbf{SYSTEM PROMPT} \\
    \texttt{You are the Chief Medical Consultant acting as the "Mediator Agent". You oversee a dialectic debate between a Proponent and an Opponent. Your job is to manage the "Consensus Graph".}
    
    \textbf{Guidelines:}
    \begin{itemize}[noitemsep, topsep=0pt]
        \item \textbf{Evaluate Validity:} Is the Opponent's counter-evidence visually grounded and logically sound?
        \item \textbf{Manage State:} Decide whether to Refute the current hypothesis or Sustain it.
        \item \textbf{Terminate:} If the debate converges or no new counter-evidence is found, declare CONSENSUS.
    \end{itemize}
    
    \vspace{0.2cm}
    \textbf{USER PROMPT} \\
    \texttt{Debate History:} \\
    \texttt{1. Proponent Hypothesis ($H_{t-1}$): \{\{OLD\_HYPOTHESIS\}\}} \\
    \texttt{2. Opponent Counter-Argument: \{\{OPPONENT\_ARGUMENT\}\}} \\
    \texttt{3. Proponent Revised Argument: \{\{PROPONENT\_RESPONSE\}\}}
    
    \textbf{Instruction:} Analyze the interaction.
    \begin{itemize}[noitemsep]
        \item Did the Proponent successfully defend their hypothesis?
        \item Or did the Opponent successfully force a revision?
        \item Is the new diagnosis consistent with all evidence seen so far?
    \end{itemize}
    
    \textbf{Output JSON:}
    \begin{verbatim}
{
  "status": "CONTINUE" or "CONSENSUS",
  "winner": "PROPONENT" or "OPPONENT",
  "current_best_diagnosis": "...",
  "confidence_score": 0.0 to 1.0,
  "explanation": "Summarize why the consensus was reached.. 
}
    \end{verbatim}
\end{agentbox}

% ...

\end{document}